\documentclass[fleqn,10pt]{wlscirep}
\usepackage[utf8]{inputenc}
\usepackage[T1]{fontenc}
\usepackage{xcolor}
\usepackage{hyperref}
\usepackage{txfonts}
\usepackage{colortbl}
\usepackage[section]{placeins}

\DeclareUnicodeCharacter{202C}{FIX ME!!!!}

\newcommand{\beginsupplement}{%
        \setcounter{table}{0}
        \renewcommand{\thetable}{S\arabic{table}}%
        \setcounter{figure}{0}
        \renewcommand{\thefigure}{S\arabic{figure}}%
     } 

\title{Beyond Size and Growth: Rethinking Lung Cancer Screening with AI‑Based Nodule Detection and Diagnosis}
\author[1,2,3,4,+]{Sylvain Bodard}
\author[5,*,+]{Pierre Baudot}
\author[5]{Benjamin Renoust}
\author[5]{Charles Voyton}
\author[5]{Gwendoline De Bie}
\author[5]{Ezequiel Geremia}
\author[5]{Van-Khoa Le}
\author[5]{Danny Francis}
\author[5]{Pierre-Henri Siot}
\author[5]{Yousra Haddou}
\author[5]{Vincent Bobin}
\author[5]{Jean-Christophe Brisset}
\author[6,7]{Carey C. Thomson}
\author[5]{Valérie Bourdès}
\author[5]{Benoit Huet}

\affil[1]{Université de Paris Cité, AP-HP, Hôpital Universitaire Necker Enfants Malades, Service d’Imagerie Adulte, F-75015, Paris, France.}
\affil[2]{Memorial Sloan Kettering Cancer Center, Department of Radiology, 1275 York Avenue, New York, NY 10065, USA.}
\affil[3]{Massachusetts General Hospital, Center for Transplantation Sciences, Harvard Medical Shool, Boston, USA.}
\affil[4]{Sorbonne Université, CNRS UMR 7371, INSERM U 1146, Laboratoire d’Imagerie Biomédicale (LIB), F-75006, Paris, France.}
\affil[5]{Median Technologies, eyonis, Valbonne, 06560, France.}
\affil[6]{Mount Auburn Hospital/Beth Israel Lahey Health, Cambridge MA, USA.}
\affil[7]{Harvard Medical School, Boston MA, USA.}

\affil[*]{Corresponding Author: pierre.baudot@mediantechnologies.com}

\affil[+]{These authors contributed equally to this work.}


\begin{abstract}
Early detection of malignant lung nodules remains constrained by size- and growth-based screening criteria, often delaying diagnosis. We present an integrated AI system that jointly performs nodule detection and malignancy assessment directly at the nodule level from low-dose CT scans, within a unified CADe/CADx framework. Unlike conventional pipelines separating detection and diagnosis, our approach targets malignant nodules directly, redefining evaluation at the point where clinical decisions are made.
To address limitations in dataset scale and explainability, the system consists in a Large Ensemble Model (LEM) combining ensembles of shallow deep learning and feature-based models and was trained and evaluated on 25,709 scans with 69,449 annotated nodules, with external validation on an independent cohort. It achieved an AUC of 0.98 internally and 0.945 externally, outperforming all growth‑based metrics, Lung-RADS\textsuperscript{\textregistered} size-based triage, European volume- and VDT-based screening criteria, radiologists and leading AI models.
The model maintains high sensitivity at low false-positive rates, excels for small and early-stage cancers, and enables malignancy assessment up to one year earlier than radiologists for indeterminate and slow-growing nodules. This approach has the potential to streamline lung cancer screening workflows and support earlier, more actionable clinical decision-making.
\end{abstract}

\begin{document}

\flushbottom

\maketitle

\thispagestyle{empty}

\section*{Introduction}

Lung cancer is the leading cause of cancer deaths worldwide\cite{barta_global_2019,siegel_cancer_2025}. Early detection through low-dose computed tomography (LDCT) dramatically improves prognosis, enabling curative treatments\cite{alberg_epidemiology_2013-1,henschke_20-year_2023,mathur_treatment_2003}, reducing mortality by over 20\% and tripling early-stage diagnoses, as demonstrated in pioneer National Lung Screening Trials such as NLST (US), NELSON (Netherlands-Belgium), and PanCan (Canada)\cite{national_lung_screening_trial_research_team_reduced_2011,de_koning_reduced_2020,tammemagi_participant_2017}.
As Lung Cancer Screening (LCS) programs are expanding globally\cite{wait_implementing_2022,parker_invitation_2023,wu_application_2024,bhamani_low-dose_2025}, new alarming trends emerged from Asian studies, with lung cancer comprising 60\% of new cancer cases\cite{lam_lung_2023}.    

Lung-RADS\textsuperscript{\textregistered} guidelines and standard care rely on nodule size as the primary malignancy predictor, with the PanCan model confirming its dominance among predictive features\cite{christensen_acr_2024-1,mcwilliams_probability_2013}. Temporal size evolution also plays a crucial role, with Volume-Doubling Time (VDT) considered the “method of choice” for malignancy assessment\cite{schwartz_biomathematical_1961,steel_growth_1966,yankelevitz_winner_2025} and currently recommended in Europe\cite{snoeckx_lung_2025,prokop_aggressiveness-guided_2025}. However, there is no consensus on its measurement\cite{christensen_acr_2024-1}and its longitudinal advantage over diameter-based assessment remains unclear\cite{lancaster_action_2025}.
Advances in quantitative imaging further reinforce the clinical importance of size-based evaluation, as highlighted by Larici et al., who stress that “size still matters”\cite{larici_lung_2017,christensen_acr_2024-1}. 

However, while morphological features contribute to malignancy prediction, their subjective nature and variability remain significant challenges\cite{nair_variable_2018,han_influence_2018}. Clinical trials from the UK and France highlight size and location as key factors in radiologist interpretation\cite{asmara_location_2024,storme_characteristics_2024}.

This continued reliance on size delays diagnosis until they meet predetermined criteria \cite{mullin_upstaging_2026-1,kim_beyond_2026-1}, i.e. size and growth thresholds and protocol-driven follow up over several months (as per NLST NELSON, LungRADs). This study focuses on the early detection and characterization of solid or part-solid parenchymal nodules (4–30 mm)\cite{bankier_fleischner_2024}, as timely intervention for these lesions is critical to improving survival outcomes\cite{gierada_survival_2021}.

AI models—particularly deep neural networks—extract high-dimensional, multi-parametric features, including morphology, that are crucial for detecting early-stage malignancies, when nodules are smallest and treatment most effective. Lung cancer is a central focus of AI-driven diagnostic research owing to the disease's global prevalence and the availability of open-access imaging data. To date, hundreds of AI models based on CT scans have been developed, with as many as 405 publications reviewed\cite{thanoon_review_2023,javed_deep_2024}. However, most models rely on a weak ground truth of radiologist assessments and suffer from an under-representation of early-stage cancer cases\cite{lancaster_action_2025}, as is the case with the LIDC dataset\cite{armato_lung_2011}. Consequently, their effectiveness is inherently limited compared to models trained on histopathology-confirmed diagnoses, such as NLST (as verified by Ma \textit{et al.}\cite{ma_novel_2023}).
Additionally, most computer-aided models are limited in scope, focusing exclusively on either nodule detection (CADe) or diagnosis (CADx). Importantly, most existing AI approaches operate at the patient level, whereas clinical decision-making in lung cancer screening fundamentally occurs at the nodule level, where detection errors directly propagate to downstream malignancy assessment. While CADe and CADx tasks are individually well studied, their integration within a single unified framework remains largely unexplored. CADe aims to detect nodules irrespective of malignancy, whereas CADx assesses malignancy only for pre-identified lesions. In contrast, an integrated CADe/CADx approach directly targets malignant nodules and assigns malignancy scores within a unified decision process, thereby redefining both error definitions and evaluation criteria.
To address this, we compare our model to Sybil\cite{mikhael_sybil_2023}, Liao \textit{et al.} (Kaggle Data Science Bowl winner; $1,972$, teams; $\$1$M prizes)\cite{liao_evaluate_2019}, and Ardila \textit{et al.}\cite{ardila_end--end_2019}—all operating at patient level, trained on NLST and with proven superior performance in lung screening populations\cite{li_no_2024}. We also benchmark against the NLST Brock-Pancan model\cite{mcwilliams_probability_2013,winter_external_2019}, recommended by the British Thoracic Society. Unlike our model, this is a purely diagnostic tool (CADx): it requires radiologists to detect nodules and assess malignancy-related features such as size, attenuation, and spiculation.
Despite AI advancements, clinical adoption of CAD systems remains hindered by high false positive (FP) rates, a major clinical concern\cite{mohammad_review_2017}. Recent studies emphasize that “increased FP nodules remain a serious drawback”\cite{fukumoto_external_2024} and stress the need to improve AI specificity\cite{geppert_software_2024}. Furthermore, models by Liao \textit{et al.}, Ardila \textit{et al.}, and Sybil provide only patient-level malignancy predictions, failing to provide explicit, directly actionable and explainable CADe/CADx-based nodule-level assessments—still lacking in published lung CAD studies at the exception of Jia \textit{et al.} recent study\cite{jia_establishing_2025}. This prevents direct comparison of detection performance at nodule level.
To address this gap, we compare our model to nnDetection, a self-configuring CAD model based on the nnU-Net design principles for detection tasks. nnDetection has demonstrated superior CADe performance, notably outperforming competitors in the LUNA16 nodule candidate-detection challenge\cite{baumgartner_nndetection_2021}.

\section*{Results}

\subsection*{Model and radiologist performances}

\begin{figure}[h!]
	\includegraphics[scale=0.63]{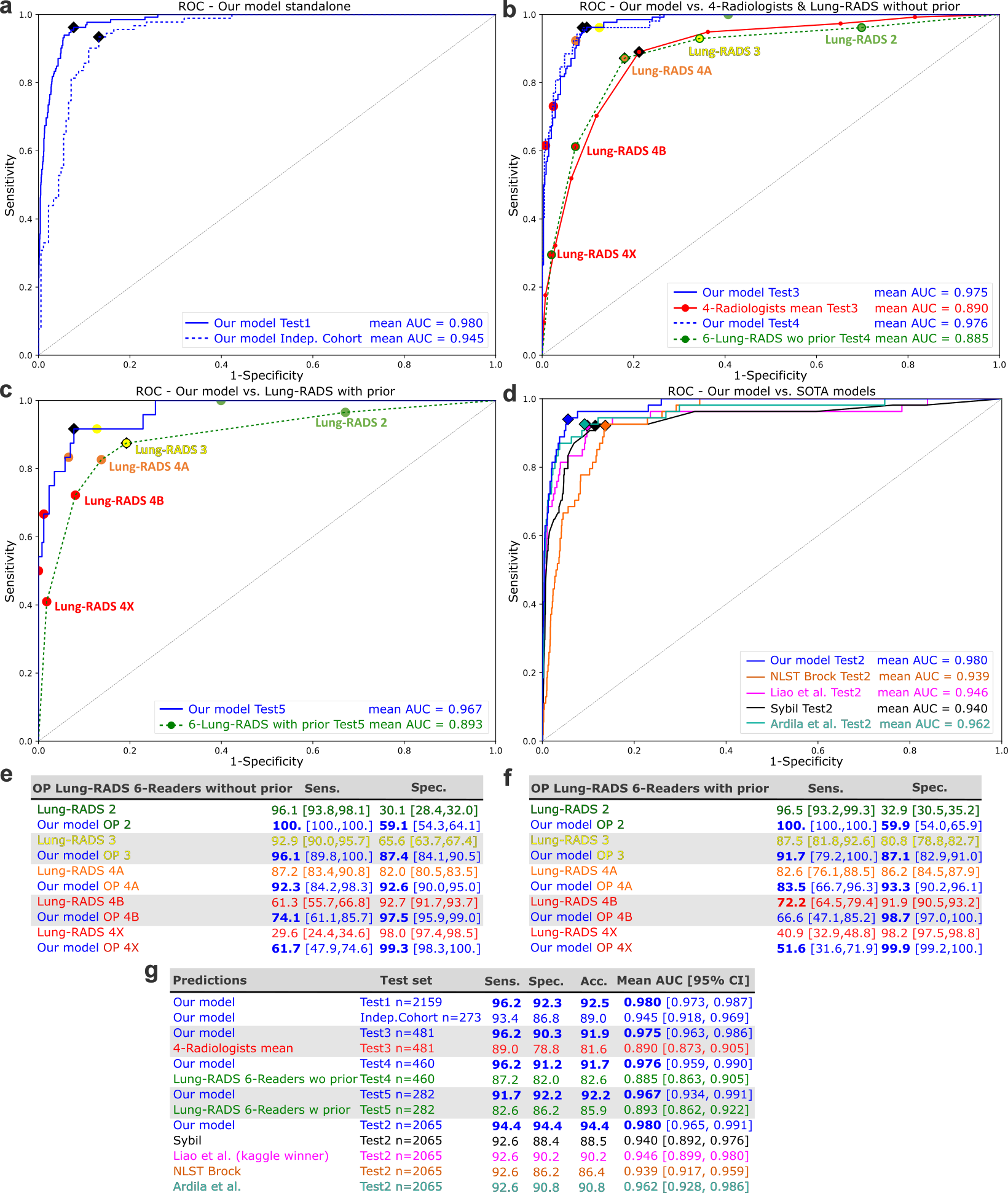}
	\centering
	\caption{\textbf{Performance comparison: our model vs. radiologists, lung-RADS\textsuperscript{\textregistered}, and SOTA models:} \textbf{a.}, Patient-level ROC curves for our model's malignancy prediction on Test1 and Independent Cohort. Our model's Operating points (OPs) at the Maximum Youden Index (MYI) are depicted with black tilted squares (in all panels of all figures except in d. where their colour  match the colour of their corresponding curve). \textbf{b.}, Patient-level ROC curves for our model's malignancy predictions on Test3 and Test4, compared with the mean of 4-Radiologists detection and Likelihood Of Malignancy assessment on Test3, as well as of six radiologists Lung-RADS\textsuperscript{\textregistered}v1.1 score assessments by six radiologists without prior Time Point (TP) evolution on Test4, as provided by Ardila \textit{et al.}\cite{ardila_end--end_2019}. Our model's  OPs equivalent to each Lung-RADS\textsuperscript{\textregistered} score are depicted with colored circles matching the Lung-RADS\textsuperscript{\textregistered}  scores on the corresponding curves. \textbf{c.}, Patient-level ROC curves for malignancy prediction comparing our model and Lung-RADS\textsuperscript{\textregistered}v1.1 score assessments by six radiologists  with prior TP evolution on Test5 as provided by Ardila \textit{et al.}\cite{ardila_end--end_2019}. Same OP representation as in \textbf{b}.  \textbf{d.}, Patient-level ROC curves on Test2 (Sybil and Ardila \textit{et al.} test set), comparing our model, Sybil (five ensembled model), Liao \textit{et al.}, Ardila \textit{et al.} models, and the NLST Brock models using NLST GT for each nodule detected by NLST radiologists. \textbf{e.}, the sensitivity and specificity with 95\% CI over 5,000 bootstraps for each Lung-RADS\textsuperscript{\textregistered} score without prior images, as assessed by six radiologists and for the corresponding accuracy equivalent OP for our model on Test4 (associated to the OPs in \textbf{b}.). \textbf{f.}, same as e., but for Lung-RADS\textsuperscript{\textregistered}  with prior on Test5 (associated to the OPs in \textbf{c}.). \textbf{g.}, Sensitivity, specificity and accuracy at the MYI of each ROC, along with the mean AUC with 95\% CI over 5,000 bootstraps.} \label{fig1}
\end{figure}

We evaluated the ability of our model to detect and characterize malignant lung nodules using two independent test sets: NLST (Test1, $n=2159$); and an Independent Cohort (IC, $n=273$). Fig.\ref{fig1} provides model evaluation through Receiver Operating Characteristic (ROC) curves, Area Under the Curve (AUC), sensitivity, specificity, and accuracy at the Maximum Youden Index (MYI). The model achieved a patient-level AUC of 0.98 on Test1 (Confidence Intervals (CIs) provided in figures). Malignancy likelihood distributions for cancer and non-cancer patients are depicted in Supplementary Fig.1.

\begin{figure}[h!]
	\includegraphics[scale=1.1]{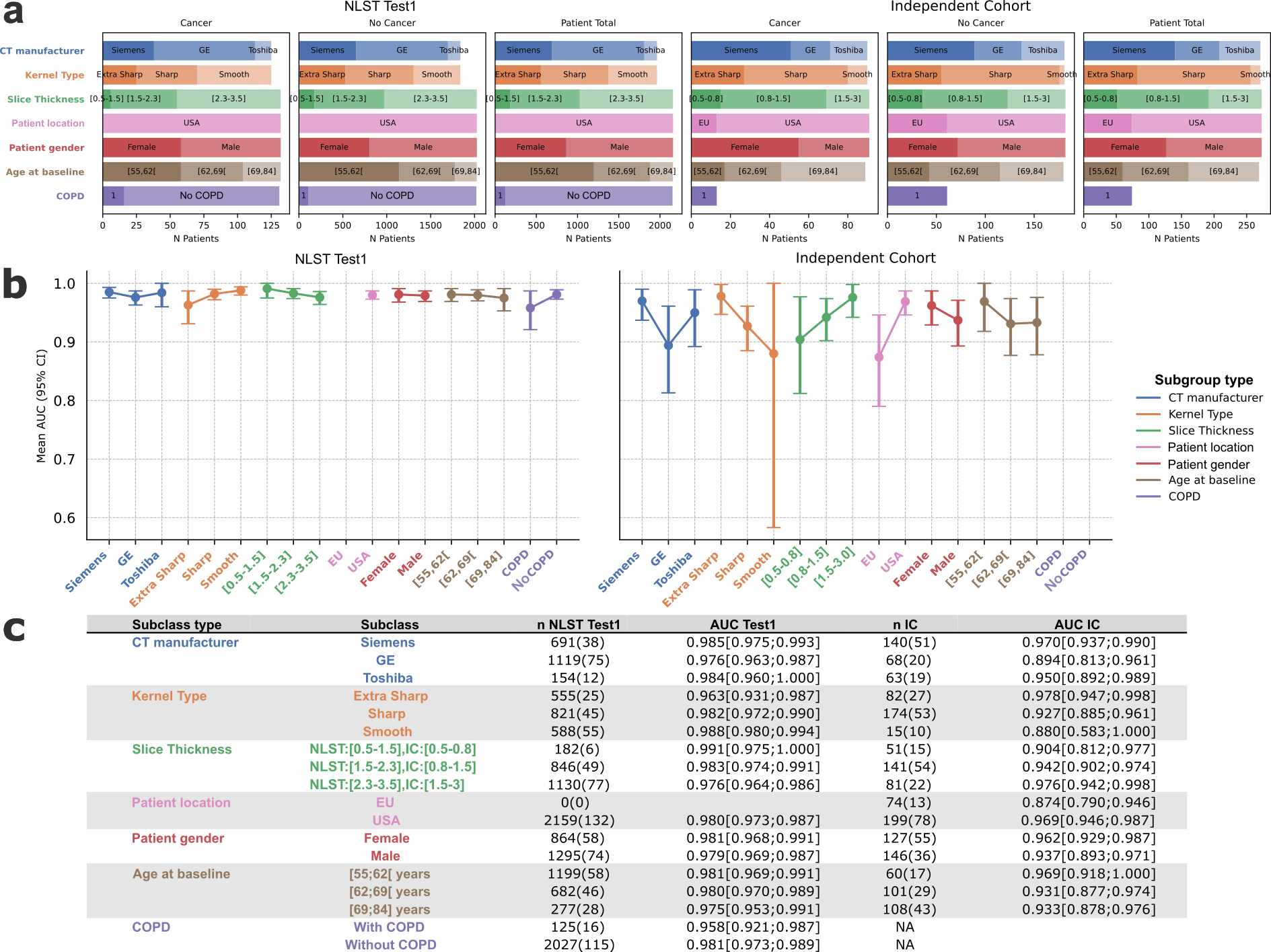}
	\centering
	\caption{\textbf{Subgroup demographics and model performance analysis}: Subgroup definitions are detailed in Supplementary methods.\textbf{a.} The distribution of demographic and scan characteristics in Test1 and IC for cancer, non-cancer and all patients. Sample sizes for Canon and Philips manufacturers are too small to be represented and are therefore replaced by white space. \textbf{b.} Mean patient AUC for the various subgroups of our model on Test1 and IC. Vertical bars represent the 95\% CI on 5,000 bootstraps. \textbf{c.} Table summarizing the values from \textbf{a.} and \textbf{b.} The sample size (n: number of patients) for each subgroup is given with the number of patients with cancer indicated in parenthesis. It provides the mean AUC and 95\% CI over 5,000 bootstraps samples. 'NA' stands for 'Not Available': the presence of COPD at baseline was only available for the EU/AIR subset of IC and is 100\% ($n=273$(91)), inclusion criterion).} \label{fig2}
\end{figure}

Validation on IC (91 cancer, 182 non-cancer) yielded an AUC of 0.945. Subgroup analysis (Fig.\ref{fig2}) demonstrated stable performance across manufacturers, kernel sharpness, slice thickness, age, and sex. However, the EU subgroup had a lower AUC (0.874) than the US subgroup (0.969), closely matching NLST (US) results. This likely reflects the EU cases originating from a Chronic Obstructive Pulmonary Disease (COPD)-only LCS campaign (AIR\cite{leroy_circulating_2017-1}), a factor known to impair model performance\cite{piskorski_malignancy_2025} as corroborated here. The lower performance observed for the GE-manufacturer scans can be explained by its over-representation in the EU/COPD subgroup (72\%, versus 8\% in IC US).\\

\begin{figure}[h!]
	\includegraphics[scale=0.9]{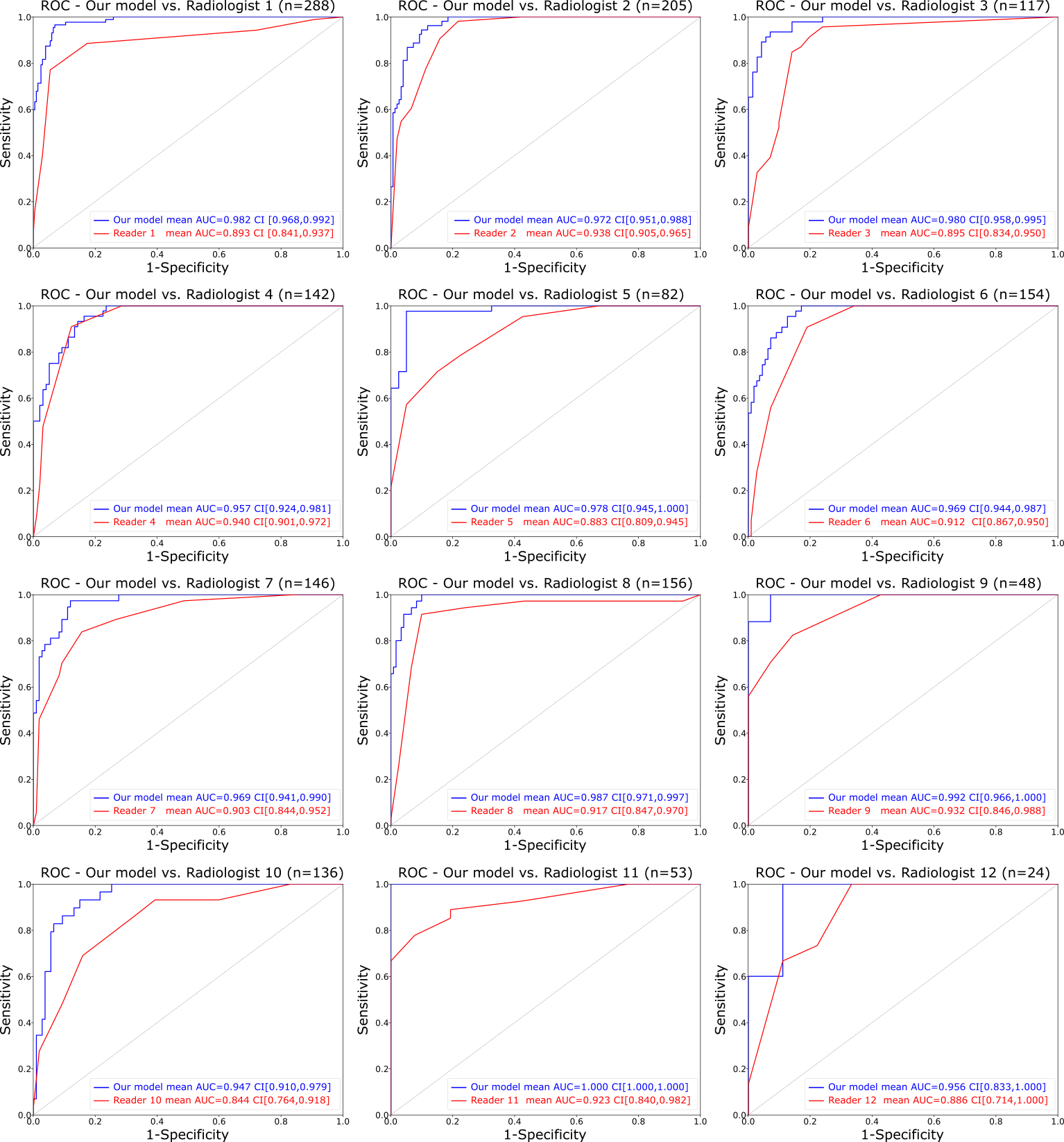}
	\centering
	\caption{\textbf{Performance comparison: our model vs. individual radiologists:} patient-level ROC curves for each of the 12 radiologist who annotated at least 15 patients with cancer and for our model on the same annotated sample of Test1, as detailed in Supplementary methods (the sample size is provided in the title of each ROC). Their mean AUC and CI over 5,000 bootstraps are indicated in the labels, and reported in detail in Supplementary Fig.2.} \label{fig3}
\end{figure}

On Test3 ($n=481$)—a subset of Test1 with four reads by 20 radiologists of varying levels of seniority, each with at least one year of experience interpreting chest CT scans (cf. Supplementary Material), hereafter "4-radiologists", assessing nodule detection and malignancy likelihood—the model achieved an AUC of 0.975, significantly surpassing the 4-radiologists' mean AUC of 0.89, ($p<0.0001$, see  Methods). Individual radiologist performance (Fig.\ref{fig3}, Supplementary Fig.2) ranged from 0.844 to 0.940 (mean: $0.906 \pm 0.028$), whereas our model consistently exceeded in AUC each of them (mean: $0.974 \pm 0.016$). Welch’s t-test confirmed non-inferiority to each individual radiologist ($p<0.0001$). Readers 13 to 20 had insufficient sample sizes for comparison.\\

On Test4 ($n=460$)—a subset of Test1 with Lung-RADS\textsuperscript{\textregistered} assessments—our model achieved an AUC of 0.976, significantly exceeding radiologists’ mean Lung-RADS\textsuperscript{\textregistered} assessment without prior CT scans (AUC=0.885, $p<0.0001$). Similar superiority was observed on Test5, where prior CT imaging was available (AUC=0.967 vs. 0.893, $p<0.0001$). As anticipated, Lung-RADS\textsuperscript{\textregistered} assessments without prior scans approximated radiologist performance, reflecting the Lung-RADS\textsuperscript{\textregistered} standardizes the methodology used by radiologists. Tables e and f of Fig.\ref{fig1} demonstrate that our model outperformed Lung-RADS\textsuperscript{\textregistered} scores without and with prior imaging in terms of specificity and sensitivity across Test4 and Test5 ($p<0.0001$), except for 4B sensitivity score with prior imaging.\\

On Test2 ($n=2065$)—a subset of Test1 that overlaps with Sybil's and Ardila et al's test sets—our model achieved an AUC of 0.980, significantly outperforming Sybil's AUC of 0.94 ($p<0.0001$)\cite{mikhael_sybil_2023}. Our model also significantly outperformed Liao \textit{et al.} (AUC=0.946, $p<0.0001$)\cite{liao_evaluate_2019}. The hitherto best state-of-the-art (SOTA) model by Ardila \textit{et al.}\cite{ardila_end--end_2019}, achieved an AUC of 0.962 on Test2, significantly below ours ($p<0.0001$). Moreover, Ardila \textit{et al.}'s model outperformed Sybil and Liao \textit{et al.}. 

Most published models focus on characterization (CADx) rather than detection (CADe), often relying on the LIDC dataset\cite{armato_lung_2011} (16 references in Ma \textit{et al.}\cite{ma_novel_2023}), where malignancy assessments are radiologist-based rather than biopsy-confirmed. Given our model's superior performance over radiologist assessments (Fig.\ref{fig1}), it inherently outperforms models trained on weak radiological Ground Truth (GT).\\

We further compared our model to the widely used NLST  Brock model\cite{mcwilliams_probability_2013,winter_external_2019}. This CADx models require radiologists to detect nodules and assess up to 11 nodule and patient-based features. Our model requires no radiologist feature assessments and significantly outperformed Brock (AUC=0.939, $p<0.0001$).\\

The performances of Sybil, Ardila \textit{et al.} and NLST Brock were slightly higher than the originally reported values (0.92\cite{mikhael_sybil_2023}, 0.959\cite{ardila_end--end_2019} and 0.912\cite{winter_external_2019}, respectively).

\subsection*{Model and radiologist detection performances}

This section details the detection performance of our integrated CADe/CADx model and demonstrates how it effectively mitigates the high FP rate challenge outlined in the Introduction.

\begin{figure}[h!]
	\includegraphics[scale=0.8]{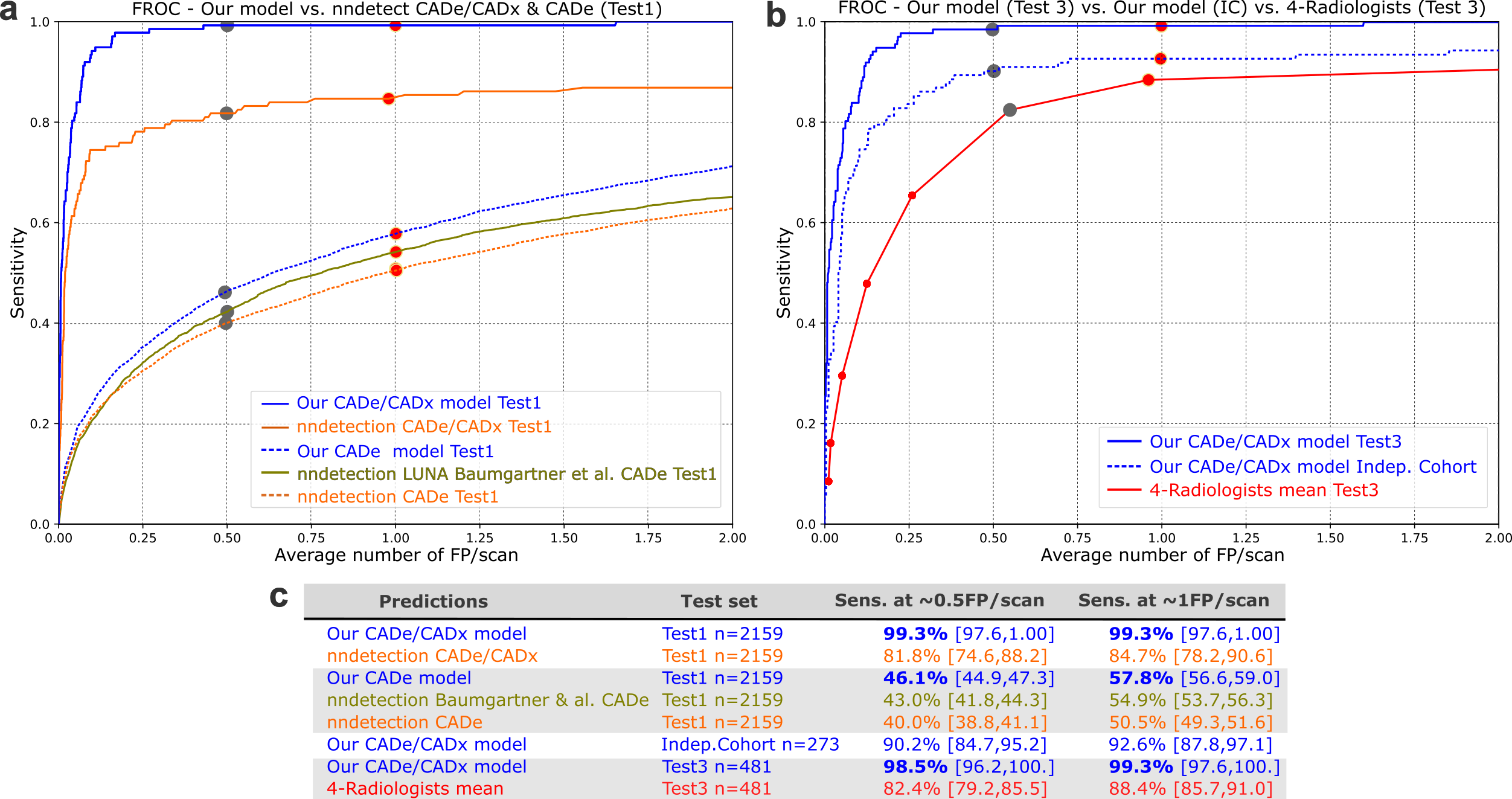}
	\centering
	\caption{\textbf{Comparison of detection performances: our model vs. radiologists, CADe and SOTA model:} \textbf{a.} The Free-response Receiver Operating Curves (FROC) on Test1 comparing our CADe/CADx model, and the nnDetection CADe/CADx model retrained on Train1 (malignant class detection only). Also shown: FROCs of our CADe-only module for the nodule detection task (malignant and benign class detection), of the nnDetection model retrained on NLST (both for malignant and benign class detection), and of the nnDetection model trained by Baumgartner \textit{\textit{et al.}}\cite{baumgartner_nndetection_2021} on LUNA16  (both for malignant and benign class detection). \textbf{b.} FROCs of our CADe/CADx model on Test3 and IC, compared to the mean performance of 4-Radiologists assessments.} \label{fig4}
\end{figure}

At the nodule level, we compared our model's performance to radiologists in detecting malignant nodules using the Free-response Receiver Operating Characteristic (FROC) curve, which measures sensitivity as a function of the mean number of FPs per scan and for Test1 and IC. Fig.\ref{fig4}a,b,c illustrates the FROCs and sensitivity values closest to 0.5 FP/scan and  1 FP/scan.

On Test1, the model achieves  99.3\% sensitivity at 0.5 FP/scan and 1 FP/scan. Performance drops slightly on the IC, with 90.2\% sensitivity at 0.5 FP/scan and 92.6\% at 1 FP/scan. For comparison, we include the FROC of the nnDetection model\cite{baumgartner_nndetection_2021}, re-trained on Train1 to detect malignant and benign nodules as separate targets (CADe/CADx). As illustrated in Fig.\ref{fig4}a, the nnDetection FROC remains below ours due to CADx module's the effectiveness in reducing FPs. On Test1, nnDetection achieved  mean sensitivities of 84.9\% at 1 FP/scan, and 82.0\% at 0.5 FP/scan—both significantly lower than our model ($p < 0.0001$).

Traditional CADe systems focus on nodule detection rather than malignancy classification, leading to significantly higher FP rates than our integrated CADe/CADx model. This disparity arises because benign nodules are considerably more prone to detection errors than malignant ones, as illustrated in Supplementary Fig.3a. 
Direct evaluations of our model's CADe component, alongside of nnDetection—retrained on NLST for separate benign and malignant detection and nnDetection trained by Baumgartner \textit{et al.}\cite{baumgartner_nndetection_2021} on LUNA16—for single-class nodule detection, further substantiate this observation (Fig.\ref{fig4}a). 
Furthermore, the nnDetection model trained on LUNA16 by Baumgartner \textit{et al.} (single-class nodule detection) outperforms the retrained nnDetection on NLST for malignant and benign nodule as separate class after joining the classes, likely reflecting the added complexity in distinguishing benign from malignant nodules within a finite model size.

Our model’s CADe component continues to outperform both nnDetection-based CADe models as a result of its large ensembling strategy while concurrently exhibiting much higher FP rates than its CADe/CADx counterparts. As expected, the performance of nnDetection on NLST is markedly lower than on LUNA16\cite{baumgartner_nndetection_2021}, since LUNA16 official evaluation includes numerous suspicious findings that are not penalized as FP if detected, thus artificially inflating performance. Accordingly, Supplementary Fig.3b shows both our CADe and LUNA16-trained nnDetection perform strongly on the LUNA16 task, matching previous reports of “outperforming all previous methods on the nodule-candidate-detection task"\cite{baumgartner_nndetection_2021}. On this set, nnDetection showed a slight performance advantage over our CADe model, likely because the former was specifically trained on LUNA16 specific task with exclusion, whereas ours was trained on NLST.\\

\begin{figure}[!htb]
	\includegraphics[scale=0.8]{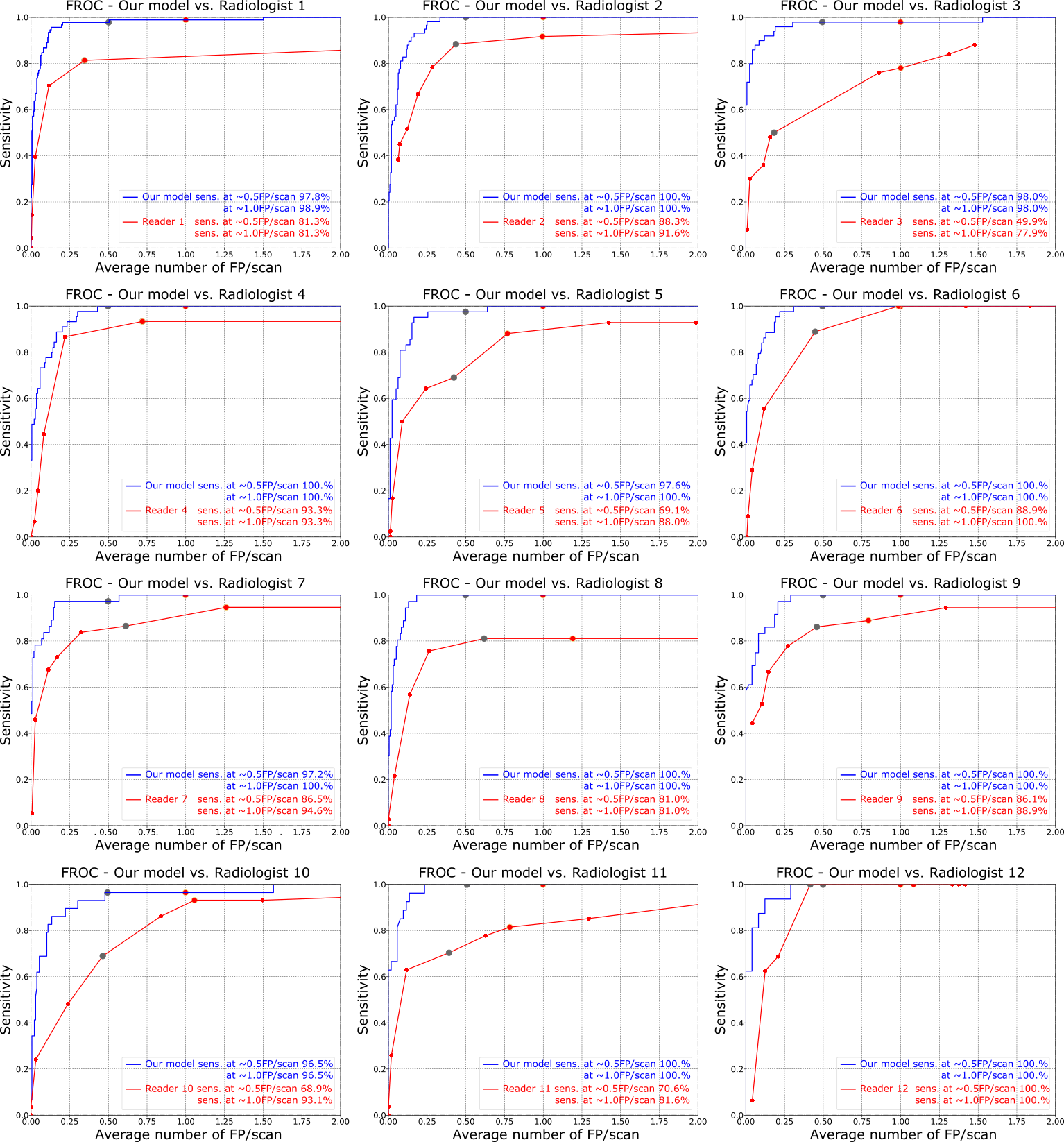}
	\centering
	\caption{\textbf{Detection performance: our model vs. individual radiologists} The 12 FROCs of our model and of the 12 radiologists who annotated at least 15 cancer patients for the same scans subsets of Test3, as detailed in Supplementary methods. Mean sensitivity over 5,000 bootstraps at the closest OP to 0.5 and 1 FP/scan is provided in labels, marked by a large grey and a large red circle, respectively.} \label{fig5}
\end{figure}

Our CADe/CADx also surpassed radiologists on the same task (Fig.\ref{fig4}b). On Test3, at 0.5 and 1 FP/scan, the mean sensitivity of 4-Radiologists was 82.4\% and 88.4\%, respectively, significantly below our model's 98.5\% and 99.3\% ($p<0.0001$). Fig.\ref{fig5} and Supplementary Fig.4 further demonstrate that our model’s FROC exceeded each radiologist performance on their annotated subsets. Radiologists failed to detect 8.2\% of malignant nodules (mean diameter $7.3$ mm$\pm4.1$ s.d), all detected by our model.\\
  
\subsection*{Beyond size with AI predictions}

\begin{figure}[!htb]
	\includegraphics[scale=0.8]{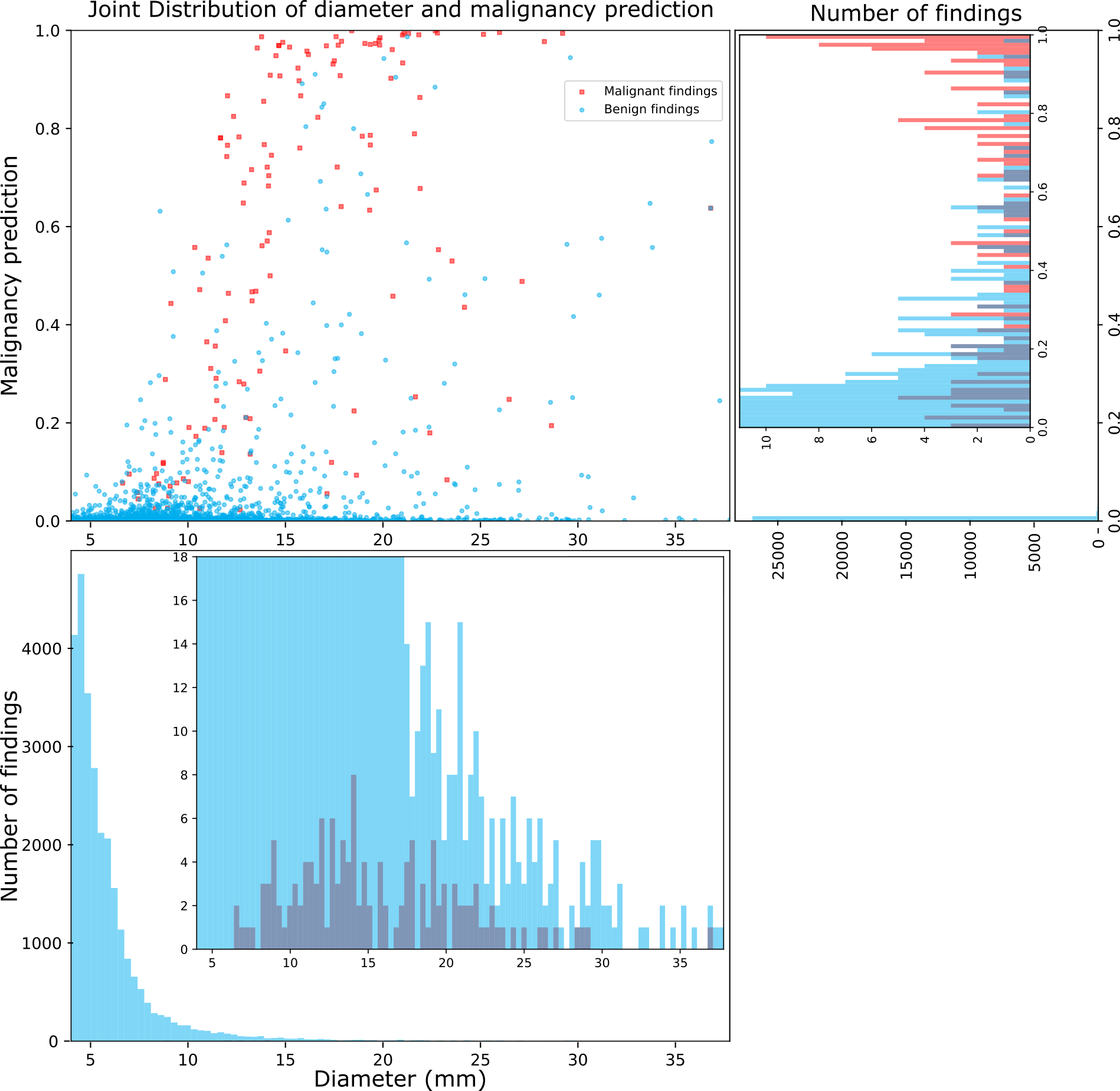}
	\centering
	\caption{\textbf{Diameter and malignancy prediction joint-distributions and marginals}: The diameter (mm) estimated from our model's automatic segmentation vs. our model's malignancy prediction for all our model detections on Test1 (with diameter greater or equal to 4 mm). The corresponding marginal distribution histograms of diameters and malignancy prediction are provided in the bottom and right panel, respectively. Due to the large imbalance between malignant and benign findings in Test1, causing cancer distributions to be nearly invisible, a zoomed-in cartoon on the distribution of the model's predictions is included in both marginal distributions.} \label{fig6}
\end{figure}

The correlation between our model’s predictions and nodule size is moderate (Pearson $\rho=0.499$), as evidenced by the joint-distribution of diameter and malignancy prediction for detections in Fig.\ref{fig6}. 

Marginal distributions of model-estimated diameter, alongside malignancy predictions for malignant and benign detections, highlight that size alone is a weak discriminator between malignancy and benignancy, with complete overlap in size but only partial overlap in malignancy likelihood. Using diameter or volume alone yields an AUC of 0.832 (CI[0.803,0.859]) and 0.876 (CI[0.844,0.906]) respectively, significantly lower than our model, corroborating the recent results of Creamer \textit{et al.} who reported an AUC of 0.822 (CI[0.787,0.857]) and 0.844 (CI[0.807,0.882]) for diameter and volume respectively on the SUMMIT Study\cite{creamer_performance_2025}.\\

\begin{figure}[h!]
	\includegraphics[scale=0.64]{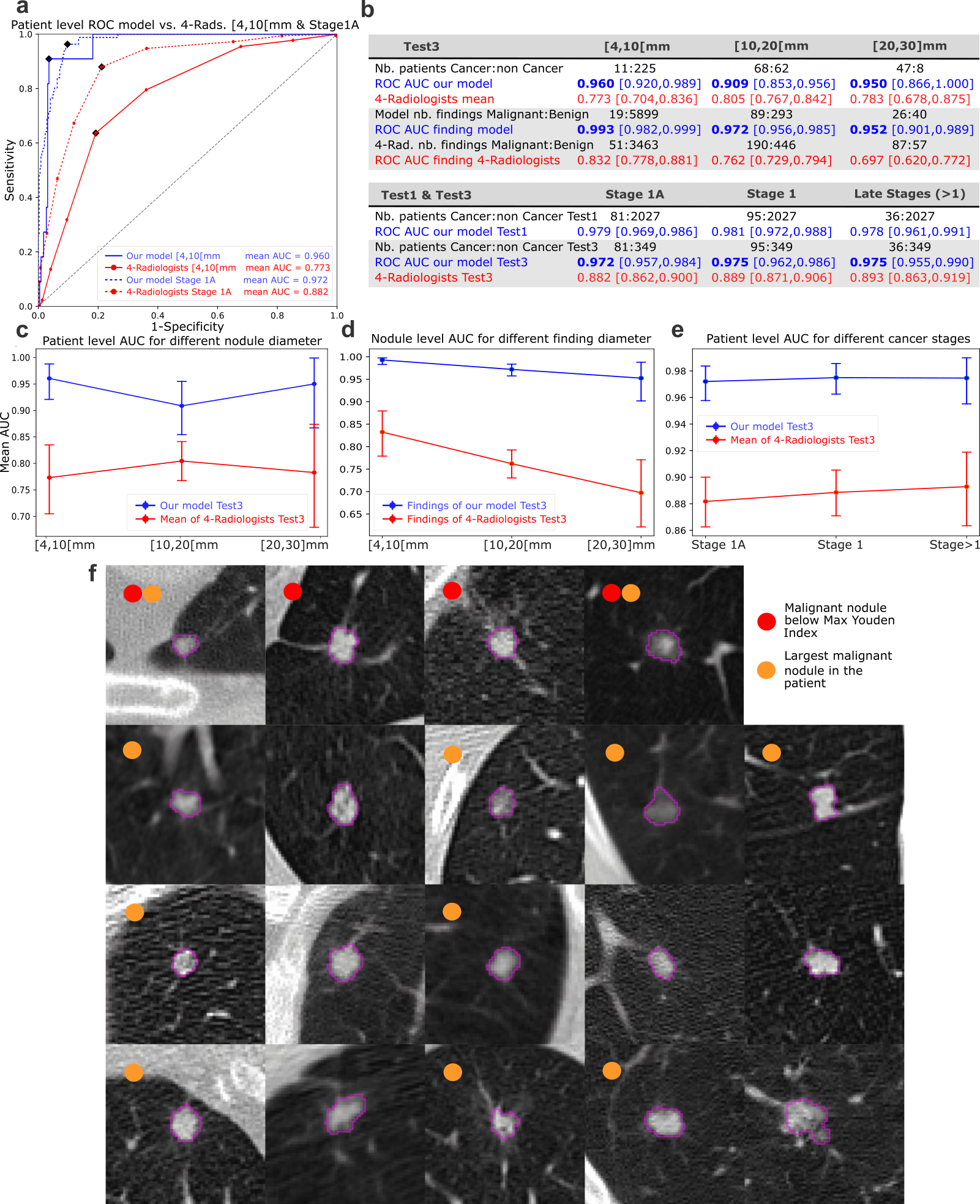}
	\centering
	\caption{\textbf{Performance across size-based subgroups:} \textbf{a.} Patient-level ROC curves on Test3 comparing our model's malignancy prediction and the mean of 4-Radiologists’ assessments for patients whose largest nodule has a diameter GT within $[4-10]$mm range, and for stage IA (vs. non-cancer) patients. \textbf{b.} Table summarizing  panels \textbf{a},\textbf{c},\textbf{d} and \textbf{e} of patient- and nodule-level AUCs for our model and for the 4-Radiologists across nodule size and cancer stage subpopulations with the corresponding malignant and benign sample size. $95\%$ are provided over 5,000 bootstraps. The discrepancy between the 19 malignant nodules (nodule level) and the 11 cancers (patient-level) reflects patients with multiple malignant nodules and differences in diameter between GT and our model estimates. \textbf{c.} Mean  AUC for subgroups of patients whose largest nodule has a diameter GT within $[4,10[$mm, $[10,20[$mm and $[20,30]$mm ranges on Test3. Horizontal bars represent the 95\% CI over 5,000 bootstraps (as for other panels). \textbf{d.} Nodule-level mean AUC of the mean of four Radiologists' findings compared to our model's findings with computed diameter in the $[4,10[$mm, $[10,20[$mm and $[20,30]$mm ranges on Test3. \textbf{e.} Mean AUC for patients with stage IA, stage 1 (stage IA and stage IB), and late stages (stage >1), compared to non-cancer patient. This stratification results from the small prevalence of stage IB and late stages in NLST (see Supplementary Information). \textbf{f.} Our model's outputs for the 19 malignant nodules in the $[4-10[$mm range in GT. Pink contours represent the automatic segmentation of the nodule. Each CT patch is 40*40mm. Nodules are ordered by ascending malignancy prediction in reading order. Red dots mark the four nodules misclassified as malignant (False Negative at Maximum Youden Index). Orange dots denote the 11 nodules identified as the patient's largest nodule of the patient (as in \textbf{b}), in all other cases, the patient presents some larger malignant nodule.} \label{fig7}
\end{figure}

At the patient level, Fig.\ref{fig7}a presents ROC curves for cancer cases with largest malignant nodule in the $[4,10[$mm range ($n=11$), and for non-cancer patients with largest nodule in the same range on Test3 ($n=225$). For these patients with small nodules, our model achieves an AUC of 0.960, significantly exceeded 4-Radiologists (AUC=0.773, $p<0.0001$). Fig.\ref{fig7}c extends this comparison across  GT diameter ranges ($[4,10[$mm, $[10,20[$mm, and $[20,30]$mm), with the model significantly in exceeding radiologists in AUC across all subgroups ($p<0.0001$).\\

Fig.\ref{fig7}d expands this analysis at nodule level on Test3, comparing malignancy detection and classification performance by the 4-Radiologists and our model. Diameters were measured from their respective segmentations. Across all diameter ranges, the model’s AUC remains stable, maintaining strong discriminative power even for small malignant nodules and significantly exceeding radiologists across all size ranges ($p<0.0001$).\\

As recommended by Lancaster \textit{et al.}\cite{lancaster_action_2025}—and supported by the cancer stage distribution in Supplementary Fig.8—NLST mainly includes early-stage cancers. To evaluate early-stage cancers performance, Fig.\ref{fig7}e presents AUC for stage IA ($n=81$, see also Fig.\ref{fig7}a), stage I (IA+IB, $n=95$) and later stages in Test1. The model achieved AUCs of 0.972 for stage IA, 0.975 for stage I, and 0.975 for later stages in Test3 (0.979, 0.981, 0.978 in Test1), significantly exceeding the mean of 4-Radiologists ($p<0.0001$).\\

Further analyses examined benign nodules in the $[4,10[$mm range that underwent unnecessary invasive procedures (e.g., biopsy-confirmed benign cases), constituting radiologist-induced FPs, with considerable patient risks and healthcare burden. In Test1, nine benign nodules meet this criterion. Using the MYI threshold, the model would have correctly classified six of these (67\%), thus holding promise for preventing unnecessary interventions.\\

Fig.\ref{fig7}f displays CT patches of all 19 malignant nodules within the $[4,10[$mm range from Test1, with their automatic segmentations generated by our model. Four nodules fall below the MYI threshold, and are thus misclassified as False Negatives at this OP. Notably, three are Solid Pleura-Attached Nodules (SPANS), a diagnostically challenging class often considered as benign\cite{snoeckx_lung_2025,zhu_how_2025}. Jiang \textit{et al.}\cite{jiang_differentiation_2023} describe SPANS as rarely detected by radiologists, particularly when small. Our model successfully detects these nodules, but underestimates malignancy likelihood for some of them.\\

Among the 15 correctly classified malignant nodules, all were solid or part-solid solitary nodules, with one exhibiting spiculation. Of these, five were SPANS, all correctly identified by the model, indicating only partial misclassifications (3/9, 33\%).\\

\subsection*{Longitudinal predictions: beyond size growth}

Nodule size growth is the key diagnostic feature, with Volume-Doubling Time (VDT) widely used to quantify changes\cite{yankelevitz_winner_2025,steel_growth_1966,arai_tumor_1994-2,nakahashi_tumor_2023}, notably in the NELSON screening protocol\cite{xu_nodule_2006}, which classifies nodules using 400- and 600-day VDT thresholds. Jennings \textit{et al.}\cite{jennings_distribution_2006-1} introduced Reciprocal Doubling Time (RDT=365/VDT), strictly equivalent to Volume Growth (\%) in AUC performance and increasing with suspicion of malignancy. Other standard measures include diameter change ($\Delta$diameter $(D_{-1}-D_{-2})$), used by Lung-RADS\textsuperscript{\textregistered}\cite{christensen_acr_2024-1}, and volume change ($\Delta$volume $(V_{-1}-V_{-2})$).

The results above summarize AI predictions at a single time point (TP): either the scan closest to cancer diagnosis within 12 months ($T_{-1}$) or the earliest available scan for non-cancer patients. To further assess predictive performance at earlier TP $T_{-2}$ (12–24 months before cancer diagnosis, when available), following Sybil and Ardila \textit{et al.}\cite{ardila_end--end_2019,mikhael_sybil_2023} or the earliest pairs of consecutive available scan for non-cancer patients, we analyzed its predictive evolution alongside standard growth measures.

\begin{figure}[h!]
	\includegraphics[scale=0.55]{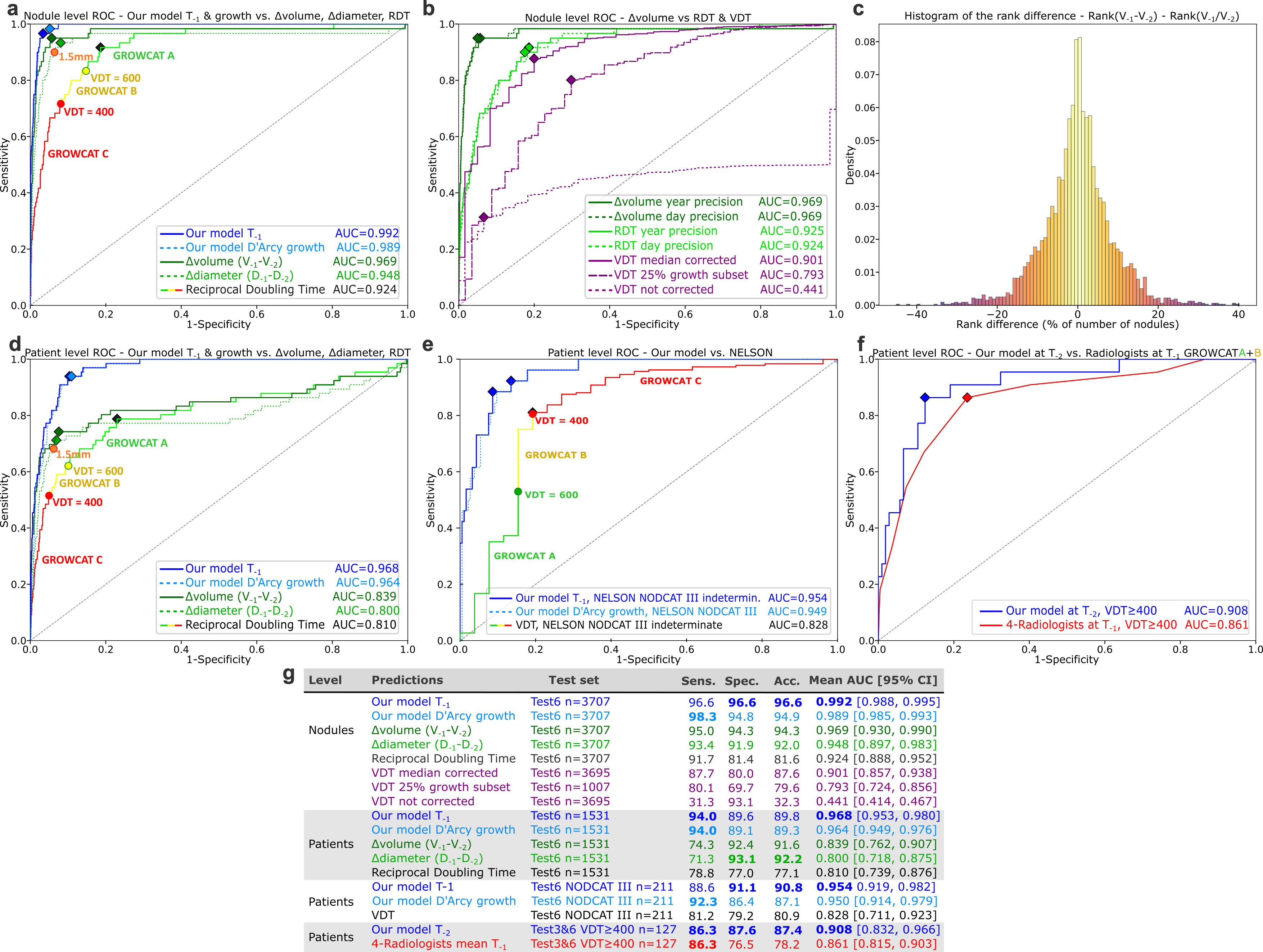}
	\centering
	\caption{\textbf{AI prediction and temporal evolution vs. standard growth metrics \& comparison of radiologists’ predictions vs. AI predictions made one year earlier (in slow-growing or indeterminate nodules)}. On each ROC the Maximum Youden Index is represented by a square matching the curve color. 
    \textbf{a.} Nodule-level ROC of  model predictions at $T_{-1}$, of the D'Arcy Thompson growth of model predictions, for $\Delta$volume, $\Delta$diameter, and RDT (or, equivalently, the percent of Volume Growth) for all nodules longitudinally paired by radiologists in GT on Test6. For RDT, the OP of 600 and 400 days thresholds—defining the GROWCAT A, B and C categories of nodule growth in the NELSON protocol\cite{xu_nodule_2006}—are depicted by yellow and red circles, while for $\Delta$diameter the threshold of 1.5 mm of Lung-RADS\textsuperscript{\textregistered}\cite{christensen_acr_2024-1} is depicted  by an orange circle. 
    \textbf{b.} Nodule-level ROC curves comparing $\Delta$volume with precision in years (same as in \textbf{a}), or precision in days ($365*(V_{-1}-V_{-2})/(T_{-1}-T_{-2})$, T in days) with RDT with precision in years, or with precision in days (same as in \textbf{a}) and VDT in three variants: without correction; with $\geq25\%$ growth selection as in NELSON~\cite{zhang_growth_2020}; and VDT with median centering correction. All VDT measures use daily precision; 12 nodules where removed with respect to \textbf{a} due to infinite VDT values (stable volume). ROC curves are computed with permuted benign/malignant GT. The median centering correction is $VDT_c=(VDT-median_{VDT})^2$ following the results and conclusion of Ho\cite{ho_effect_2017}. The VDT with $\geq25\%$ growth selection yielded an AUC of 0.795, exceeding but comparable to the 0.67 reported by~\cite{zhang_growth_2020} in different population.
    \textbf{c.} Distribution of rank difference—expressed as a percentage of the total number of nodules—between the $\Delta$volume and the volume ratio (as for RDT) over the nodules longitudinally paired in GT of Test6. 
    \textbf{d.} Patient-level ROC of the same variables as in \textbf{a} with the same representations, obtained by selecting the largest nodule at first TP for size growth analysis, per the NELSON protocol, using our model's patient-level predictions for the curves and its temporal evolution. The sample of 1,531 patients includes all Test6 patients with at least one nodule longitudinally paired by radiologists in GT.
    \textbf{e.} Performance comparison of the NELSON protocol using NELSON criteria\cite{xu_nodule_2006,klaveren_management_2009} and our model on NLST: Patient-level ROC curves for our model and VDT on the subgroup of patients with nodule with volume at  $T_{-2}$ in the $[50-500]mm^3$ range and showing at least a 25\% volume increase (NODCAT III indeterminate category).
    \textbf{f.} Patient-level ROC comparing our model at $T_{-2}$ vs. the mean of 4-Radiologist at $T_{-1}$ for the subgroup of patients of Test6 with nodules with VDT$\geq$400 days, i.e., corresponding to the negative and indeterminate growth category in the NELSON protocol\cite{xu_nodule_2006} (GROWCAT A+B). 
    \textbf{g.} Mean sensitivity, specificity and accuracy at the MYI, sample size and mean AUC with 95\% CI over 5,000 bootstraps for each ROC presented in \textbf{a,b,c,d,e,f}.} \label{fig8}
\end{figure}

Fig.\ref{fig8}a compares the ROC curves of our model’s "instantaneous" prediction at $T_{-1}$, the evolution of its prediction (using D'Arcy Thompson’s growth function\cite{bonner_growth_1945,reed_growth_1919}p.160), and three standard growth measures ($\Delta$volume, $\Delta$diameter, RDT) for all nodules longitudinally paired by radiologists in GT on Test6.

The best-performing function for prediction evolution follows D’Arcy Thompson’s autocalytic model (logistic) of biological growth: $P_{-1}/(P_{-1}-P_{-2})_n$, both outperforming simple ratios or $\Delta$-based  functions. Surprisingly, this function remains marginally inferior to the instantaneous prediction at $T_{-1}$. Our model’s instantaneous and evolution predictions significantly outperform all standard size growth measures ($p<0.0001$), with  AUCs of 0.992 and 0.989, respectively. Supplementary Fig.5 illustrates that the growth of model prediction yields less overlap between benign and malignant nodule distributions than traditional metrics. RDT performs worst with an AUC of 0.924, whereas $\Delta$volume, as expected, slightly outperforms $\Delta$diameter (AUC=0.969), confirming that accurately measured size growth is the primary indicator of malignancy\cite{yankelevitz_winner_2025}. Fig.\ref{fig8}b shows VDT’s non-linearity requires either median-centering correction\cite{ho_effect_2017}, or pre-selecting nodules with $\geq25\%$ growth, as in NELSON\cite{zhang_growth_2020}. Even after median correction, VDT performed weakest (AUC=0.901); dropping to 0.441 uncorrected. Limiting to nodules with $\geq25\%$ growth boosts AUC to 0.795, exceeding the 0.67 reported by~\cite{zhang_growth_2020}.
Fig.\ref{fig8}b reveals that normalizing VDT or $\Delta$volume to daily precision exerts negligible non-significant effects on AUC compared to year-precision normalization used for $\Delta$volume or $\Delta$diameter ($p>0.1$). However, Fig.\ref{fig8}c reveals ranking differences of up to 40\% between $\Delta$ and volume ratios, e.g., a nodule ranked first over 1,000 by volume ratio may rank 400th by $\Delta$volume (-40\%). This ranking discrepancy explains the performance gap between RDT/VDT and $\Delta$volume.\\

To assess clinical impact, we computed the same ROC curves (Fig.\ref{fig8}d) following the NELSON protocol, which prioritizes the largest nodule at the initial TP for size growth analysis and our model's patient-level predictions. Patient-level results mirror with nodule-level findings—except RDT marginally outperforming $\Delta$diameter, reflecting size selection errors—while the gap between AI predictions and standard measures widens (AUC=0.969 for AI vs. 0.81 for RDT). Delta‑volume significantly outperformed conventional growth measures ($p<0.0001$) and showed an optimal threshold of 72 mm$^3$/year (MYI). 

Our RDT estimates for malignant nodules strongly correlate with Jennings \textit{et al.}\cite{jennings_distribution_2006-1} (Pearson $\rho=0.908$, $p<0.0001$; Supplementary Fig.6), confirming consistency with prior work. Fig.\ref{fig8}e quantifies NELSON's protocol\cite{klaveren_management_2009} on NLST, selecting indeterminate (50-500 mm$^3$) and growing ($\geq$25\% Volume Growth) NODCAT III nodules at initial TP ($T_{-2}$) and evaluating VDT over short follow-up. In this subgroup, our model achieved an AUC of 0.954, improving sensitivity by 11.9\% and specificity by 7.4\% over NELSON’s VDT criteria at their respective MYI. These findings also validate NELSON’s 400-day VDT and 1.5 mm $\Delta$diameter screening thresholds (Fig.\ref{fig8}ad) aligning closely with their MYI on ROC curves.

Using $T_{-2}$ predictions, we compared early AI diagnosis to that of 4-radiologists on the follow-up scan acquired one year later at $T_{-1}$. As anticipated, for NELSON’s fastest-growing nodules (GROWCAT C, VDT<400 days), radiologists outperformed AI at $T_{-2}$ but were subsequently surpassed by it at $T_{-1}$ (Supplementary Fig.7). In contrast, Fig.\ref{fig8}f demonstrates that for slow-growing and indeterminate nodules (GROWCAT A+B, VDT$\geq$400 days), our model at $T_{-2}$ is statistically equivalent to radiologists at $T_{-1}$ ($p<0.0001$, AUC=0.908 for AI vs. 0.861 for 4-radiologists).

\section*{Discussion}

Previous studies have demonstrated AI models can equal or surpass radiologists in diagnostic accuracy across various medical imaging tasks\cite{rodriguez-ruiz_stand-alone_2019, wu_deep_2020, mckinney_international_2020, pacile_improving_2020-1}. In LCS, Ardila \textit{et al.} reported superior AI performance\cite{ardila_end--end_2019}, and the Brock model improved triage over Lung-RADS\textsuperscript{\textregistered}v1.1\cite{mcwilliams_pl0214_2024}. Our study confirms these observations, and presents an architecture that significantly outperforms existing models, including those by Ardila \textit{et al.}\cite{ardila_end--end_2019}, NLST Brock\cite{winter_external_2019}, Liao \textit{et al.}\cite{liao_evaluate_2019} (winner of the Kaggle Data Science Bowl), and Sybil\cite{mikhael_sybil_2023}. However, Liao and Brock performances may be inflated in our comparison due to possible training–Test1 overlap and resulting overfitting.\\
In contrast, our model exceeds the mean performance in AUC of 4-radiologists and each of individual radiologist, as well as Lung-RADS\textsuperscript{\textregistered} assessment with or without prior imaging—whereas previous studies only matched Lung-RADS\textsuperscript{\textregistered} when prior scans were available\cite{ardila_end--end_2019}. Additionally, our model consistently surpasses radiologist performance across all nodule sizes and cancer stage subgroups. However, by design, our model is limited to solid and part-solid parenchymal nodules [4,30] mm, and is therefore not intended to detect hilar, mediastinal, Ground Glass Opacity (GGO) nodules, or incidental findings, which still rely entirely on the reading of radiologists.
Interestingly, for cancers with slow or indeterminate growth ($VDT\geq400$ days), our model achieves comparable performance $[24,12]$ months earlier than radiologists evaluating follow-up scans acquired one year later (AUC=0.908 vs. 0.861, $p<0.0001$). This suggests the model’s potential for detecting malignancy well before overt metastatic progression, enabling earlier intervention and improved patient outcomes. \\
Our model’s performance was further confirmed in an independent cohort, providing supportive evidence of external validity. However, this study remains retrospective, and although the relatively small sample size limits the robustness of subgroup analyses, these findings should be strengthened through larger prospective evaluations in real-world screening settings.\\

Our model also achieves a detection sensitivity of 99.3\% at 0.5 FP/scan, a substantial improvement over SOTA detection models, far lower than traditional CADe systems, which typically focus on detecting all nodules without assessing malignancy. For example, the DL-LND\cite{fukumoto_external_2024} reported a sensitivity of 93.0\% at 7 FP/scan, and Murchison \textit{et al.} achieved 95.9\% sensitivity at 10.9 FP/scan\cite{murchison_validation_2022}. Moreover, many of our model’s FP are benign nodules that conventional CADe models would classify as true positives, highlighting the advantage of an integrated CADe/CADx approach. Indeed, while conventional CADe models aim to detect all nodules, our model selectively targets suspicious ones, complicating direct comparisons. Since benign nodules are considerably more challenging to detect (cf. supplementary Fig.3a)—due to ambiguity in GT assessments and inter-reader variability\cite{armato_assessment_2009}—traditional CADe systems generally exhibit poor specificity. In contrast, malignant nodule detection benefits from more reliable GT, enabling deep learning models to achieve higher accuracy. Clinically, this distinction is critical: CADx-supported malignancy ranking improves screening efficiency by prioritizing the most relevant nodules. 

This gain is primarily driven by the effectiveness of the CADx module in filtering out FPs, powered by a massively parallel ensemble of highly diverse classifiers—including 2D and 3D CNNs and tree-based models—that collectively bolster its performance and generalization. Unlike nnDetection, Sybil, or Ardila—which rely on end-to-end backpropagation—our model adopts a factorized learning strategy, that separates detection ('where') from characterization ('what') within a frozen learning framework. This design partially mirrors the dual-stream architecture of human visual cortical processing, enhancing explainability, by automatically linking malignancy suspicion to precise nodule localization and detailed features like diameter, volume, and radiomics. This strategy facilitates more efficient learning, particularly with medium-sized datasets. 
Three factors underlie our model’s superior performance. First, its architecture—comparable in size to ViT-Huge—adopts a fundamentally different approach using a modular, factorized, and large-scale ensembling of specialized shallow modules, optimized for mid-sized medical datasets. Second, it leverages a highly accurate GT, from biopsy-confirmed diagnoses and follow-up data, complemented by detailed internal annotations. Third, the training dataset includes over 23,000 CT scans from LIDC and NLST, making it one of the largest publicly available lung cancer AI datasets, second only to  Chen \textit{et al.}\cite{chen_cancerunit_2023} 
Generic large multi-modal transformers remain hard to scale for 3D imaging due to their high data volume and computational demands. Even Med-Gemini\cite{yang_advancing_2024,saab_capabilities_2024} or MedMPT (0.824 AUC on NLST\cite{ma_visionlanguage_2025}) still far underperform specialized models like ours. Scarcity of curated, high-quality training data—not methodology—is now the main bottleneck. Under current constraints, specialized shallow models ensemble remains the most effective solution.


Radiologists rely heavily on nodule size and growth dynamics for malignancy triage, as formalized in current lung cancer screening workflows such as Lung‑RADS\textsuperscript{\textregistered} in the United States and volume‑ and VDT‑based criteria in the European NELSON protocol. In contrast, our AI model demonstrates higher accuracy and robustness than standard size‑based measures, and a single AI‑based malignancy prediction at $T_{-1}$ outperforms all size‑derived growth metrics, supporting the prioritization of AI‑driven risk assessment over traditional size‑ and growth‑based approaches. In the absence of AI, $\Delta$Volume is identified in our analysis as the strongest size‑derived predictor, outperforming $\Delta$diameter and VDT, and suggesting its potential relevance for future growth‑based assessment strategies.
By surpassing Lung‑RADS\textsuperscript{\textregistered}–based triage and NELSON volume‑ and VDT‑based triage, our approach suggests that AI‑based malignancy prediction could reduce unnecessary follow‑up imaging and invasive procedures—major contributors to patient anxiety and healthcare costs—while improving early cancer detection. The integration of such AI capabilities into clinical workflows therefore holds substantial promise for transforming lung cancer screening paradigms on both sides of the Atlantic. Furthermore, early identification of malignancies among small nodules that are currently managed conservatively may drive changes in clinical decision‑making, enabling earlier intervention and potentially improved outcomes\cite{loh_ct-guided_2013}. Collectively, these findings support a re‑evaluation of existing screening workflows in light of AI‑enabled CADe/CADx systems.

In conclusion, our findings set a strong reference point for AI-assisted LCS. The model significantly exceeds radiologists' performances in both malignant nodule detection and cancer diagnostic—despite using a single time point, unlike radiologists who rely on longitudinal scans—challenging the limitations of size-based assessments and redefining what is possible in screening workflows. Future work should prioritize the real-world integration of AI tools to enhance early diagnosis, reduce unnecessary interventions, and ease the healthcare burden. Importantly, the broader economic benefits of AI-enabled earlier detection and treatment deserve further investigation to inform policy and reimbursement strategies.

\section*{Methods}

\subsection*{Data}
This study was designed and reported in accordance with established guidelines for predictive modeling and diagnostic accuracy studies, including TRIPOD and STARD-AI, where applicable. The dataset consists of three cohorts: LIDC-IDRI, NLST, and an IC. All data were used in accordance with relevant ethical regulations. Publicly available datasets (LIDC-IDRI and NLST) are de-identified and were accessed in compliance with their respective data use agreements. For the independent cohort, ethical approval was obtained from the relevant institutional review boards, and all data were handled in accordance with applicable regulations and patient privacy requirements.        

\subsubsection*{Datasets} 

\paragraph*{LIDC-IDRI:} A publicly available US LCS database with 1,018 CT scans from 1,010 patients\cite{armato_lung_2011}. It provides high-quality nodule detection, segmentation, and malignancy assessments from up to eight annotators. However, histopathological confirmation is available only for 77 malignant and 36 benign cases. Further details are provided in Supplementary methods.

\paragraph*{NLST:} The National Lung Screening Trial is a large, multi‑centric U.S. screening program with longitudinal follow‑up extending up to seven years\cite{jemal_lung_2017}. CT scans were acquired across 33 institutions, providing a fair sampling of the heterogeneity encountered in real-world lung cancer screening populations, with biopsy‑confirmed cancer diagnoses but no nodule‑level segmentations or localizations. To address this, we enriched its GT via two re-annotation campaigns:
\begin{itemize}
\item Complementing GT (Test1, Train1) with full nodule detection, longitudinal pairing, segmentation, and characterization.
\item Assessing radiologist performance using a protocol similar to LIDC, with four annotations per scan from 20 radiologists (Test3).
\end{itemize}
For further details, see Supplementary methods.

\paragraph*{Independent Cohort:} A multi-centric dataset from three providers:
\begin{itemize}
\item Segmed (US, $n=255$, 49\%)
\item Gradient (US, $n=131$, 25.2\%)
\item AIR Trial\cite{leroy_circulating_2017-1} (France, $n=134$, 25.8\%), comprising exclusively patients with COPD.
\end{itemize} 
Further details are in Supplementary methods. Demographic and scan parameter distributions for NLST and the IC are presented in Fig.\ref{fig2}.

\subsubsection*{Data inclusion/exclusion}

\begin{figure}[h!]
	\includegraphics[scale=0.17]{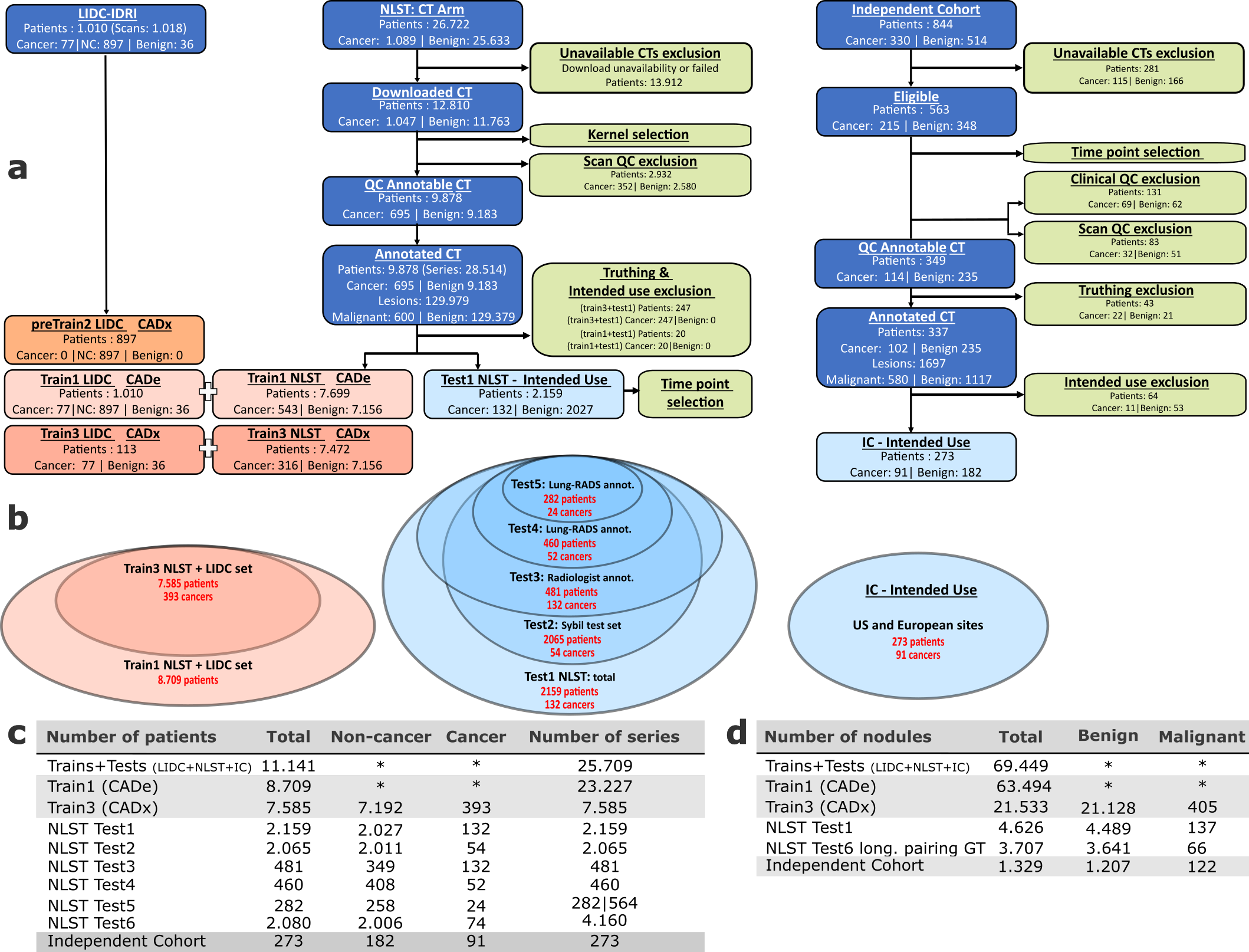}
	\centering
	\caption{\textbf{Datasets inclusion/exclusion:} \textbf{a.} Study inclusion–exclusion flowchart with sample sizes. For LIDC train sets, 'Cancer' refers to cases confirmed by histopathology study following biopsy or resection; 'Non-cancer' refers to cases confirmed negative for cancer by at least one year of follow-up; 'not confirmed' includes all other cases (for which a malignancy visual assessment based solely on CT is given by multiple annotators) \textbf{b.} Venn diagrams representing inclusions of our train and test sets. \textbf{c,d.} Patient, series and nodule counts for each of our train and test sets. * denotes values that are not applicable due to lack of definitive cancer GT in full LIDC datasets.} \label{fig9}
\end{figure}

A complete overview of the dataset inclusion/exclusion flow is illustrated in Fig.\ref{fig9}. 
Fig.\ref{fig9} a. depicts the different stages of the inclusion/exclusion process of the study, alongside the corresponding sample sizes. 
All steps of inclusion/exclusion process and corresponding criteria are described in the Supplementary methods. For NLST, we adhered as closely as possible to the inclusion/exclusion workflow and criteria of Ardila \textit{et al.}\cite{ardila_end--end_2019}, considering our specific nodule level malignancy detection aims. This enables direct comparison by ensuring our patient and volume datasets closely approximate those used in Ardila \textit{et al.} (and Sybil\cite{mikhael_sybil_2023}). Following classical nodule-level study inclusion (see for example \cite{baldwin_external_2020}), our deviations from Ardila \textit{et al.} protocol are that we exclude cancer patients presenting only malignant (pure) GGO nodules—focusing on both solid and part-solid—owing to the higher malignancy risk associated with part-solid nodules\cite{naidich_recommendations_2013} and the comparatively lower likelihood of malignancy for pure GGO (see\cite{li_diagnosis_2024} and references therein), and that, implementing the definition of nodule of the Fleischner society \cite{bankier_fleischner_2024}, we focus only on malignant nodule in the [4,30]mm diameter range while malignant nodules above 30 mm—considered as masses—are excluded (notably mediastinal and hilar masses). GGO display atypical slow growth and excellent survival outcomes, may unlikely progress to malignancy only after becoming part-solid \cite{korb_elusive_2018,herskovitz_detection_2022}, and have diameter and volume inherently fuzzily defined, inducing considerable measurement variability\cite{goodman_inherent_2006}, rendering them unsuitable for the purpose of this study.
As a result of this selection, 2,065 patients (Test2) overlap both the 2,357 and 2,203 patients of the respective test sets of Ardila \textit{et al.}\cite{mikhael_sybil_2023} and of Sybil\cite{mikhael_sybil_2023}, and 481 (Test3) overlap the 501 annotated test set of Ardila \textit{et al.} \\

For NLST and IC, the selection consists of three main steps with inclusion/exclusion criteria specified (see Supplementary methods for details): 
\begin{enumerate}
    \item Availability: scans available without download error;
    \item Quality control (QC): Exclusion of kernels, patients and scans lacking complete information (e.g. three TPs for NLST) or unreadable images,
    \item Eligibility: exclusion based on study purpose and truthing exclusion. This step excludes all non-parenchymal cancers (hilar, mediastinal), patients with exclusively ground-glass malignant nodules, and patients whose malignant nodules fall outside the diameter ranges of $[3,40]$mm and $[4,30]$mm for training and test sets, respectively. Additionally, scans from cancer patients that could not be disambiguated—where our radiologists were unable to identify the malignant nodule on the available CT scan—are also excluded .
    \item Time Point selection: for Test1–5, the scan closest to cancer diagnosis within 12 months ($T_{-1}$) or, for non-cancer patients, the earliest available scan. For Test6 (longitudinal), for cancer patients the same TP scan as test1–5 for $T_{-1}$ and the scan closest to cancer diagnosis within $[12,24]$ months for $T_{-2}$; for non-cancer patients, the earliest pair of consecutive TPs available.
\end{enumerate} 
For LIDC, which serves solely as a training dataset and is already curated, we only included cases confirmed by biopsy or follow-up (see Supplementary methods for details).\\

\subsubsection*{Train and test sets} \label{NLST test sets}

Following the inclusion/exclusion process, the NLST-LIDC dataset was split at the patient level into two subsets—a training set and a test set—by “leave-out” such that: (1) they are fully independent; (2) the test set ($n=2,159$) has maximal overlap with the test set in Ardila \textit{et al.}\cite{ardila_end--end_2019} obtained from random splitting, and hence supposed to preserve covariate distributions of intended use population ; and (3). the test comprises 132 cancer patients, i.e., the minimum sample size determined by a preliminary statistical power analysis to ensure robust overall and subgroup analyses. As the test set of Ardila \textit{et al.}\cite{ardila_end--end_2019} contained only 66 cancer cases after exclusions, we supplemented it with 66 additional cancer patients randomly selected from NLST to enrich our test set (Test1). \\

\paragraph*{Training Sets:}
\begin{itemize}
\item \textbf{Train1:} used for CADe training, includes all TPs, excludes pure GGO cancers, and retains only nodules of interest (see Supplementary methods);
\item \textbf{Pretrain2:} used for CADx pretraining, includes LIDC-IDRI scans excluded from Train3 (i.e., without confirmed cancer or non-cancer status);
\item \textbf{Train3:} used for CADx training, includes one single TP per patient, excludes pure GGO cancers, and retains nodules in the $[3,40]$mm range.
\end{itemize}
Train1 and Train3 were subsequently split into training and tuning subsets for CADe and CADx training, respectively, as detailed in Supplementary methods.
Fig.\ref{fig9}c. reports the sample sizes as well as the cancer-non-cancer breakdown and class imbalance across the training and test sets. 

\paragraph*{Test Sets:}
\begin{itemize}
\item \textbf{Test1} ($n=2,159$; 132 cancers): the largest test set, reflecting real-world LCS conditions, derived from the extensive multi-centre NLST campaign and enriched to 6.1\% cancer prevalence to support more robust subgroup analyses;
\item \textbf{Test2} ($n=2,065$; 54 cancers): Subset of Test1 used for benchmarking against SOTA models. It maximally overlaps with the test sets of Sybil\cite{mikhael_sybil_2023} and Ardila \textit{et al.}\cite{ardila_end--end_2019}, thereby mitigating overfitting bias. This set exhibits an imbalance (2.6\% cancer prevalence) that better approximates the expected LCS population.
\item \textbf{Test3} ($n=481$; 132 cancers): Subset of Test1, enriched in cancers (28\%) for robust subgroup analysis. Annotated by 4-Radiologists for nodule detection and malignancy scoring, it maximizes overlap with the 507-patient test set from Ardila \textit{et al.}\cite{ardila_end--end_2019}.
\item \textbf{Test4} ($n=460$; 52 cancers): Subset of Test1 originating from Ardila \textit{et al.}\cite{ardila_end--end_2019}, featuring Lung-RADS\textsuperscript{\textregistered} assessments by six radiologists without access to prior CT scan information;
\item \textbf{Test5} ($n=282$; 24 cancers): Subset of Test1 derived from Ardila \textit{et al.}\cite{ardila_end--end_2019}, featuring Lung-RADS\textsuperscript{\textregistered} assessments by six radiologists, with access to prior CT scan information;
\item \textbf{Test6} ($n=2080$; 74 cancers): the earliest pair of consecutive TP scans available for patients from Test1. Two subsets of Test6 were used for longitudinal analysis: one with all (patients with) nodules longitudinally paired by radiologists in GT ($n=1531$, 66 cancers, $n=3707$ nodules, 66 malignant); the second intersecting with Test3 ($n=419$, 74 cancers). 
\end{itemize}
Fig.\ref{fig9}c,d summarizes training and test set composition.

\subsection*{Model}

\subsubsection*{Large Ensemble Model (LEM) and Strength in Specialization: A Factorized Ensemble of Small Models to Address Data and Explainability Constraints}

Medical datasets are typically smaller than those used for training Large Language Models and involve handling high-dimensional 3D medical images. This makes very deep Convolutional Neural Networks (CNNs) and transformers inefficient due to their large number of parameters and high GPU memory requirements\cite{pu_advantages_2024}. Transformers generalize CNNs by replacing spatially limited convolutional attention with exhaustive attention, thereby increasing learnable parameters, particularly in 3D\cite{dosovitskiy_image_2020}, hence requiring far larger datasets for effective training. In theory, with unlimited data and computational capacity, an optimal approach might involve a single-voxel self-attention classifier taking the entire CT scan as input or 3D CNNs in more limited settings, with detection and segmentation naturally emerging as sub-tasks. However, in data-constrained settings, 'weak learners' (e.g., gradient boosted trees), shallow networks, and ensemble methods often outperform deep models\cite{grinsztajn_why_2022,mohammed_comprehensive_2023}. Studies have shown that smaller networks like EfficientNet-b0 (45K parameters) can outperform larger counterparts in CT classification\cite{yang_comparative_2021}.
To address the challenges of training on a mid-sized dataset (\~10,000 cases), we designed a factorized ensembling strategy that combines diverse and complementary CNNs in a massively parallel architecture (Large Ensemble Model, LEM) to enhance robustness and performance. Drawing inspiration from the ventral and dorsal visual pathways in the human cortex\cite{milner_two_2008}, we factor learning into separate detection (localization) and characterization tasks—an approach successfully adopted by top-performing Kaggle teams.
Our model employs a hybrid ensemble architecture that integrates a detection module—built from 3D-SegResnet ensembles and 3D-retina UNet for lung and nodule segmentation—alongside a characterization module that combines 15 2D-CNNs, 15 3D-CNNs, 15 radiomics-based gradient boosted trees (leveraging 122 radiomics and 3D-morphomics features\cite{munoz_3d-morphomics_2022}), and a full-CT scan 3D classifier adapted from a Sybil-inspired attention model\cite{mikhael_sybil_2023}. This diverse ensemble effectively balances local and global feature extraction, for strong performance on mid-sized medical datasets.

Each component of this 'factorized' model uses a SOTA approach: a Sybil-inspired\cite{mikhael_sybil_2023} full-CT scan model captures not only local nodule features, but also contextual features (e.g., emphysema, nodule density, and location); a gradient boosted trees classifier leverages standard radiomics\cite{aerts_decoding_2014} and 3D geometrical curvature features\cite{munoz_3d-morphomics_2022}, as in recent studies\cite{xie_differential_2024,wang_value_2024,kim_clinical_2024}; and a relatively shallow 39-layer 3D DenseNet builds on prior success of similar CNNs in medical imaging studies\cite{astaraki_benign-malignant_2021}.
Additionally, this factorization strategy into subtasks also aligns with the expected standards for explainability in the medical field\cite{fillon_key_2024,hantel_perspectives_2024}: the model not only predicts cancer diagnosis but also, through the intermediate module outputs, provides detailed information on each detected finding that contributed to the final prediction. This includes precise localization of findings ranked by malignancy suspicion, their segmentation, and standard characteristics such as size, volume, and other radiomic features. However, explainability remains partial, and the performance gains offered by AI over feature-based models\cite{tammemagi_participant_2017,winter_external_2019} is distributed throughout the entire 3D scan voxels and millions of parameters.

Ensembling these diverse and complementary models with targeted pretraining modules enhances both performance and generalizability. Importantly, as confirmed by a preliminary ablation study, even simpler classifiers contribute modestly yet meaningfully to the final predictions. As detailed in Supplementary Methods (Section “Nodule prediction ensembling”), analysis of sub-module contributions indicates that 3D-CNN, 2D-CNN, and morpho-radiomics classifiers contribute approximately 92.4\%, 1.8\%, and 5.8\%, respectively, underscoring the complementary value of hybrid modeling strategies in data-constrained clinical settings. Although our model has a high total parameter count as a result of this parallel structure (965 million parameters, see Fig.\ref{fig10}), its design fundamentally differs from traditional large CNN and transformer models—in both structure and training strategy—notably by introducing systematic sub-network freezing and integrating models trained with substantially different and specialized loss, reflecting a heavily expert- and experience-driven supervised 'Frankenstein modeling' paradigm. For comparison, classic CNNs such as 2D VGG-16, ResNet-50, DenseNet169, and Inception-v4 have 138M, 23M, 7M, and 41M parameters, respectively\cite{yang_comparative_2021}, while the original ViT-Huge model (2D) has 632M parameters and SOTA large-generic medical model MedGemma-only “approaching the performance of task-specific models”-has 4B or 27B parameters\cite{sellergren_medgemma_2025}. Our model is designed for a commercial fully automated assistant for LCS radiologists.

\subsubsection*{Our model}
 
The model was developed incrementally, by integrating the ensemble of 3D-CNN, the ensemble of 2D-CNN, an XGBoost classifier based on radiomics and morphomics, and full-CT scan model—each addition validated to significantly improve classification performance, as measured by AUC on tuning set. The modules were trained sequentially in feed-forward order, using outputs from preceding modules as input, with a basic version of sub-network freezing method\cite{brock_freezeout_2017-1} that enables the use of different losses. 
\begin{figure}[h!]
	\includegraphics[scale=0.23]{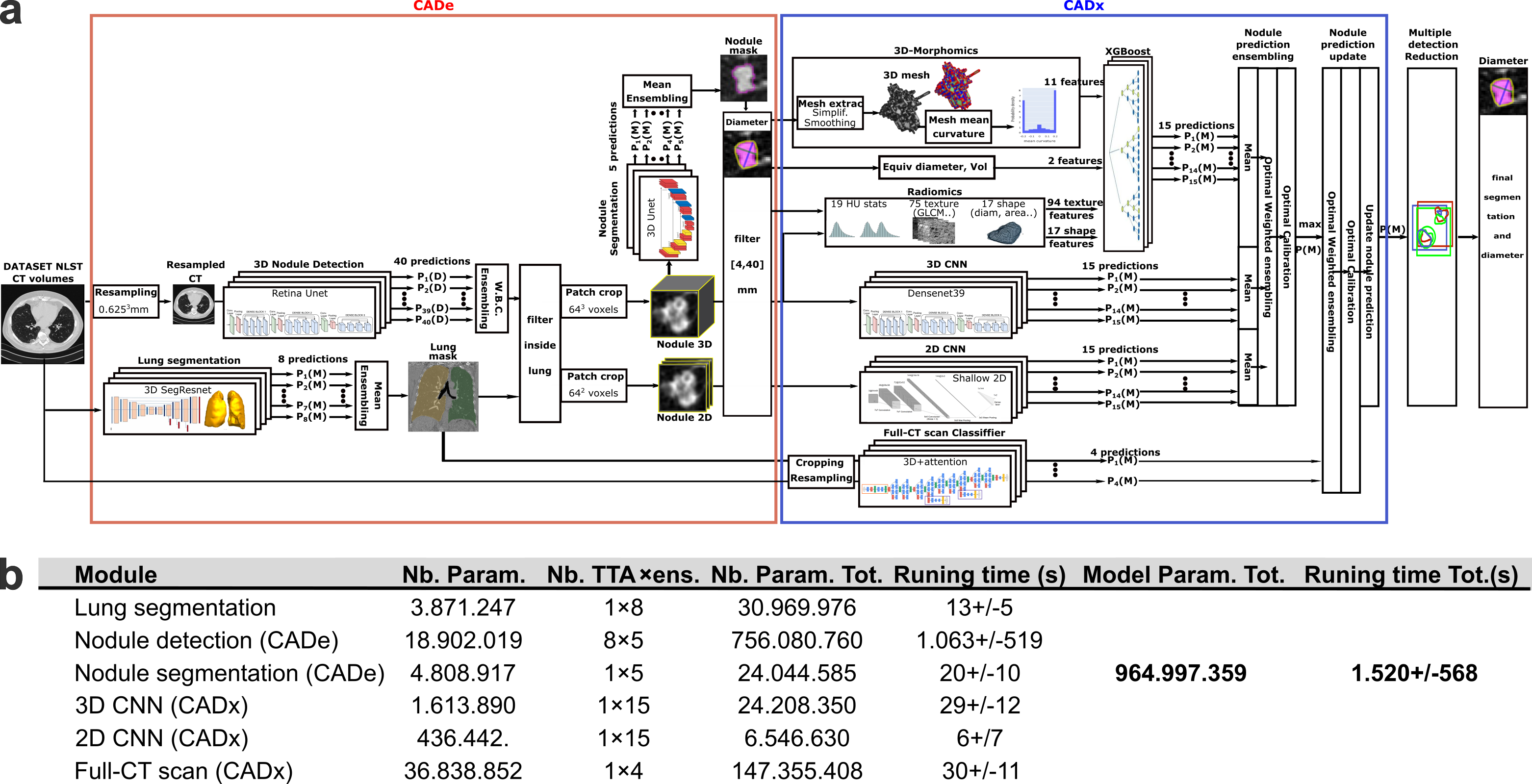}
	\centering
	\caption{\textbf{Modular architecture of the CADe/CADx:} \textbf{a.} The model comprises two sequential components: a CADe module for nodule detection (left, red box); and a CADx module for malignancy characterization (right, blue box). The CADe module includes: an ensemble of 3D CNNs for lung segmentation; another for nodule detection with FP reduction; and a third for nodule segmentation. The CADx module combines: an ensemble of gradient-boosted decision tree classifiers using radiomic and morphomic features;
    ensembles of local patch-based 2D and 3D CNNs; and a full-CT scan 3D CNN with an attention mechanism. Predictions at the nodule- and patient-levels are ensembled and integrated via calibrated stacking. A final post-processing step removes spurious and duplicate detections. \textbf{b.} Number of parameters and running time in nominal configuration (in seconds) per Deep Neural Network submodules: Number of learnable parameters of each individual model, number of Test Time Augmentation (TTA) and number of ensembled models (ens.), the total number of parameters for each Deep Neural Network modules of our model, the mean +/- stdev running time in seconds over Test1 for each Deep Neural Network modules of our model. The two last column gives the total number of parameters of the whole model and the total mean +/- stdev running time of the model.} \label{fig10}
\end{figure}

The sequential modules that make up the CADe/CADx are detailed below, and illustrated in Fig.\ref{fig10}a (from left to right): 

\begin{itemize}
\item \textbf{Lung segmentation:} ensemble of 3D-SegResNet models\cite{ronneberger_u-net_2015}.
\item \textbf{Nodule detection (CADe):} ensemble of Five Retina U-net\cite{jaeger_retina_2018-2} models with eight test-time augmentation.
\item \textbf{Patch extraction and filter in lung (CADe):}  filters out any detection outside the lung parenchyma-with Bounding boxes having Intersection over Union (IoU)>0 with the lung mask.
\item \textbf{Nodule segmentation (CADe):} ensemble of 3D-U-Net\cite{ronneberger_u-net_2015}.
\item \textbf{Diameter measurement (CADe):} extracts minimum, maximum, and mean diameters from the convex hull of the nodule segmentation. It filters out any detection outside the $[4,40]$mm diameter range.
\item \textbf{Nodule characterization (CADx):} ensemble of a 15 shallow 3D-CNN densenet39\cite{huang_densely_2017-1} module, a 15 shallow 2D CNN module, and a 15 XGBoost\cite{chen_xgboost_2016} classifiers module based on 111 extracted radiomic and 11 3D-morphomic features\cite{munoz_3d-morphomics_2022}.
\item \textbf{Nodule Predictions ensembling and calibration (CADx):} averages outputs from the 15 3D, 2D and XGBoost models, then computes the optimal weights ensembling (convex optimal stacking\cite{mohammed_comprehensive_2023}) to combine them into a single prediction per finding. It then applies optimal calibration using the NetCal logistic model by Küppers et al\cite{kuppers_multivariate_2020}.
\item \textbf{Full-CT scan characterization (CADx):} ensemble of 4 3D-CNN module with trained attention mechanism adapted from Sybil\cite{mikhael_sybil_2023} and retrained on Train3, taking the full CT scan in input.
\item \textbf{Nodule prediction update and calibration (CADx):} computes the optimal weights ensembling (convex stacking\cite{mohammed_comprehensive_2023}) of the four full-CT scan predictions alongside the maximum nodule predictions; and computes the optimally calibrated transformation of this patient prediction\cite{kuppers_multivariate_2020}. Nodule predictions are then updated by adding a correction factor proportional to the difference (in percentage) between the final patient prediction and the maximum of detected nodule predictions. As a result of this last step, the patient and maximum nodule predictions are equal, calibrated (prediction value optimally reflects the real observed accuracy), and bounded within the $[0,1]$ interval.
\item \textbf{Multiple detections reduction module and diameter measurement (CADx):} first, the central connected component of each finding is selected. Overlapping central components from different findings are merged by union, with the merged finding inheriting the highest nodule prediction. Diameters and volume are then extracted from the merged findings. For this step, the previous diameter estimation method is slightly adapted to better reflect radiological practice.
\end{itemize}
A detailed description of each of those modules, of their training and parameters is provided in the Supplementary methods.

Given the highly parallelized, modular architecture, training and inference time depend on the available computational resources. All deep neural network (DNN) modules were trained on NVIDIA A100 GPUs with 40GB of memory (a single GPU was sufficient for training).
For inference, the model was deployed using a containerized Kubernetes environment, with configurable hardware allocations depending on the deployment mode. In the nominal configuration, inference was distributed across three nodes: a CPU‑only node (4 cores, 8GB RAM), a GPU‑enabled node (2 CPU cores, 16GB RAM, NVIDIA GPU with 16GB memory), and a high‑CPU node (8–16 CPU cores, 32GB RAM).
Inference runtime varies with the number of CT slices and detected nodules per scan. Fig.\ref{fig10} reports the mean and standard deviation of inference time in nominal configuration for each DNN module, as well as the total end‑to‑end computation time.

\subsubsection*{Other models}\label{met_other_model}

\paragraph{Sybil:} The original Sybil model\cite{mikhael_sybil_2023} (\href{https://github.com/reginabarzilaygroup/Sybil}{https://github.com/reginabarzilaygroup/Sybil}), using the predictions of an ensemble of five models, is run in test mode on our Test2 using the default configuration, one year before diagnosis (which provided the best
performance for Sybil and corresponds to our model’s settings).\\

\paragraph{Liao \textit{et al.}:} The original model of Liao \textit{et al.}\cite{liao_evaluate_2019} (\href{https://github.com/lfz/DSB2017}{https://github.com/lfz/DSB2017}) is run in test mode on our Test2 using the default configuration.\\

\paragraph{Ardila \textit{et al.}:} We use the original predictions of Ardila \textit{et al.}\cite{ardila_end--end_2019} for patients on Test2 provided in their supplementary files and using the selection method described in Supplementary Information. \\

\paragraph{NLST Brock model:} The NLST Brock model, adapted from the Pancan-Brock model and recalibrated for NLST dataset complemented with a pack-year feature (thereby addressing low PPV and overestimation issues in the original model), was re-implemented following\cite{winter_external_2019} and applied to Test2 using nodule features provided by NLST radiologists GT (available at \href{https://cdas.cancer.gov/datasets/nlst/}{https://cdas.cancer.gov/datasets/nlst/}). \\

\paragraph{nnDetection model:} The nnDetection model\cite{baumgartner_nndetection_2021} (\href{https://github.com/MIC-DKFZ/nnDetection}{https://github.com/MIC-DKFZ/nnDetection}) was retrained on Train3 for CADe/CADx and Train1 for CADe to separately detect malignant nodules (CADe/CADx) or malignant and benign nodules (CADe) using the default configuration. We also tested the "original" CADe model trained on LUNA16 provided by Baumgartner\cite{baumgartner_nndetection_2021}.\\

\subsection*{Data analysis}\label{Data analysis}

\subsubsection*{ROCs and FROCs} 
Malignancy likelihood predictions for each patient is given by the highest malignancy prediction value among all detected findings for models operating at nodule level (such as ours, the Brock model and the radiologists' readings), and from the output of Sybil, Ardila \textit{et al.} and Liao \textit{et al.} (that only provides a global patient prediction). We report the MYI, which indicates the optimal cutoff threshold for a deterministic binary decision between predicted malignant and benign outcomes (cf.\cite{kallner_formulas_2018} p.106): $J=Max_c(Sensitivity_c+Specificity_c - 1)$, where $c$ is the optimal cutoff\cite{kallner_formulas_2018}p.107.
ROC curves are computed using standard Scikit learn function, with AUCs estimated via trapezoidal (linear interpolation) integration rule and averaged over 5,000 bootstrap samples. Note that Lung-RADS\textsuperscript{\textregistered} discretizes malignancy scores into only six classes, which introduces an uncertainty on the ROC and AUC estimations due to interpolation.\\
\textbf{Radiologists ROC:} the mean ROC curve and its corresponding AUC across the four sets of annotations per scan (from a total of 20 radiologists) is obtained assuming equivalence of readers (permutation invariance). In practice, this simply involves pooling the whole set of radiologist predictions against GT, essentially generalizing the non-parametric ROC method\cite{chen_average_2014-1,fawcett_introduction_2006} while allowing for varying sample sizes per annotator. To evaluate individual radiologist performance relative to our model and assess inter-radiologist variability, we select the 12 radiologists who annotated at least 15 cancer cases. We compute the ROC curves for each of these as well as our model using the same set of patients and scans for comparison. The resulting 12 AUCs and corresponding CIs are derived from 5,000 bootstraps from both individual readers and the model, using identical samples. Results are summarized in Fig.\ref{fig3}.\\ 
\textbf{ROC curves for subgroups:}
For patient-level subgroups focusing specifically on patient characteristics, we simply consider the associated patient subsets and compute their ROC. 
For diameter-based patient subgroups, we define three ranges—$[4,10[$, $[10,20[$, $[20,30]$mm—based on the largest nodule per patient. 
For nodule-level diameter subgroups analysis, findings from either radiologists or our model are selected based on their respective diameter estimations.

\subsubsection*{Statistical tests} 
For AUC, accuracy, sensitivity and specificity, the estimations of the mean, standard deviation, 95\% CI and distributions are obtained by computing 5,000 bootstrap samples with replacement\cite{efron_introduction_1994-1}. For these measures, performance comparisons between our model and others were conducted over the same bootstrap samples by applying a superiority t-test with unequal variance using scipy.stats (one-sided Welch t-test, our prediction vs. other). A superiority test assuming equal variance (t-test) produced consistent results at equivalent p levels across all cases. 

\subsubsection*{Lung-RADS\textsuperscript{\textregistered} scores equivalent OPs of the model}
The OPs of our model, designed to correspond to Lung-RADS\textsuperscript{\textregistered}  scores, follow the methodology of Ardila \textit{et al.}\cite{ardila_end--end_2019}, with one key modification: equivalence is based on accuracy rather than precision, because our model's predictions are calibrated to reflect empirical accuracy\cite{kuppers_multivariate_2020}. The accuracies of Lung-RADS\textsuperscript{\textregistered}  accuracy scores are derived from NLST estimates reported by Pinsky (See Table 2)\cite{pinsky_performance_2015} by uniformly distributing the ambiguous categories—i.e., classes 3 or 4A or 4B—across included classes (["Lung-RADS\textsuperscript{\textregistered} 1": 0.010499583,"Lung-RADS\textsuperscript{\textregistered} 2": 0.565082253,"Lung-RADS\textsuperscript{\textregistered} 3": 0.870543047, "Lung-RADS\textsuperscript{\textregistered} 4A":0.934665051, "Lung-RADS\textsuperscript{\textregistered} 4B":0.972519143, "Lung-RADS\textsuperscript{\textregistered} 4X": 0.983397773]) OPs in our training set with accuracy values closest to these Lung-RADS\textsuperscript{\textregistered} accuracies are then selected as the equivalent OPs.

\subsubsection*{Size and size evolution measures} 
\textbf{Diameter:} In line with Fleischner Society guidelines for radiological best practices for measuring pulmonary nodules\cite{bankier_recommendations_2017} and Lung-RADS\textsuperscript{\textregistered} 2022\cite{christensen_acr_2024-1}, the diameter of a finding is defined as the mean of the long- and short-axis diameter of the segmentation mask, either annotated by radiologists or automatically generated by our model using the algorithm detailed in Supplementary Information.\\

\textbf{Volume:} Nodule volume (in $mm^3$) computed from the segmentation mask provided by either the radiologist or by the model.\\

\textbf{Cancer stages:} AJCC6\cite{greene_lung_2002} stages provided by NLST.\\

\textbf{VDT, RDT and Volume Growth:} Calculated following established methods\cite{jennings_distribution_2006}, using the Schwartz formula\cite{schwartz_biomathematical_1961} $VDT=((t_{-1}-t_{-2})\ln 2)/
(\ln(V_{-1}/V_{-2}))$ and $RDT=365/VDT$, where $t_{-1}$ and $V_{-1}$ are time (in days) and volume measured at the final CT; and $t_{-2}$ and $V_{-2}$ are measured at initial CT. Volume Growth is given by\cite{xu_nodule_2006} $100\times(V_{-1}-V_{-2})/V_{-1}$. \\

\textbf{D'Arcy Thompson growth of model prediction:} given by the formula $P_{-1}/(P_{-1}-P_{-2})_n$. The logarithm is omitted here, as its monotonicity leaves the ROC invariant. The term $(P_{-1}-P_{-2})$ is normalized to the $[0,1]$ interval to prevent issues with negative values.\\

\subsubsection*{Nodule and detection pairing} 
Pairing between GT and detected nodules (be it by models or radiologists) is addressed using an IoU criteria of IoU>0.1 between GT and 3D Bounding Boxes, as recommended for 3D contexts\cite{jaeger_retina_2018-2}. 

This study follows, where applicable, established reporting guidelines for predictive modeling and diagnostic accuracy studies, including TRIPOD and STARD-AI.

\section*{Acknowledgements}

The authors gratefully acknowledge the National Cancer Institute and the Foundation for the National Institutes of Health for establishing and maintaining open access to the NLST and LIDC/IDRI Databases, which proved essential to this study. We also thank the DKFZ team, especially M. Baumgartner for sharing their code and LUNA16 trained nnDetection model. 

\section*{Funding}
Not applicable

\section*{Author contributions}

S.B., P.B., C.V., B.R., B.H. conceived the experiments, P.B., G.D.B, E.G., V.K.L, D.F., B.H. conceived the model, E.G., B.R., C.V., V.B., Y.H., P.B., V.K.L, G.D.B, D.F., carried out data acquisition and preprocessing, P.B., G.D.B, E.G., V.K.L, D.F., P.H.S., Y.H., C.V., B.R., B.H. analysed the results, S.B., P.B., B.H. wrote the manuscript, J-C.B., V.B.,  B.R., C.C.T., provided analysis suggestion and manuscript corrections. All authors reviewed the manuscript. 

\section*{Competing interests}
The authors declare the following competing interests: Median Technologies has filed for patent protections (application numbers: US11810299B1, EU22315156.4, CN 202380053824.2, and US12272063B2, EU22315034.3, CN202380028642X, JP024-548634) on behalf of B.H., D.F., P.B. for the work related to 2D and 3D CNN ensembling and on behalf of B.H., P.B., E.M., E.G., J-C.B., V.G. for the work related to 3D-morphomics. P.B., B.R., C.H., G.D.B, E.G., V.K.L, D.F., P.H.S., Y.H., J-C.B., V.B., B.H. are employees of Median Technologies and own Median Technologies stock as part of the standard compensation package.

\section*{Data availability}
LIDC-IDRI and NLST are  open-source dataset available at \href{https://www.cancerimagingarchive.net/collection/lidc-idri/}{https://www.cancerimagingarchive.net/collection/lidc-idri/}\cite{armato_lung_2011}, and \href{https://www.cancerimagingarchive.net/collection/nlst/}{https://www.cancerimagingarchive.net/collection/nlst/}\cite{national_lung_screening_trial_research_team_data_2013}. The IC is subject to restrictions and cannot be publicly released by the providers. 
The Multi-Reader annotation dataset, with a total of 2,004 scan annotations performed by radiologists—comprising 23,099 detected and annotated lesions, among which the 6,505 most suspicious nodules (implementing LIDC inclusions criteria for nodules) are segmented with malignancy assessment, for 501 patients (including those from Test3) of the public NLST dataset— is shared as open source, thereby supplementing the public LIDC annotations on NLST scans, is shared in open-source in The Cancer Imaging Archive (TCIA) under the short name LIDC-annot-NLST501, final release in progress). 
 
\section*{Code availability}
To enable automatic and full reproducibility of all figures, statistical analyses, and results presented in the article, to support further benchmarking of the reported performances, and to generically evaluate arbitrary CADe/CADx, we publicly release the evaluation chain at \href{https://github.com/EYONIS-AIDS-DS/CADe-CADx-evaluation}{https://github.com/EYONIS-AIDS-DS/CADe/CADx-evaluation}, together with the corresponding model predictions, nodule volumes, and diameters (...). The code used for the implementation of the model has dependencies on internal tooling and infrastructure, and as such, is protected by patent applications filed under the following numbers: US11810299B1, EU22315156.4, CN 202380053824.2, and US12272063B2, EU22315034.3, CN202380028642X, JP024-548634), preventing public release. In line with reproducibility and transparency standards in AI emphasized by Haibe-Kains \textit{et al.}\cite{haibe-kains_transparency_2020}, all experiments and model implementations are described thoroughly in the Methods and Supplementary Information (Methods) sections to facilitate replication with non-proprietary libraries. All relevant links to the necessary library repositories are provided. To promote responsible innovation, we are actively seeking collaboration with research partners, regulators, and healthcare providers to evaluate and research secure applications and support further model advancement and improvement.  


\clearpage

\section*{Supplementary Information}
\beginsupplement

\section*{Supplementary Figures}

\begin{figure}[!htb]
	\includegraphics[scale=0.4]{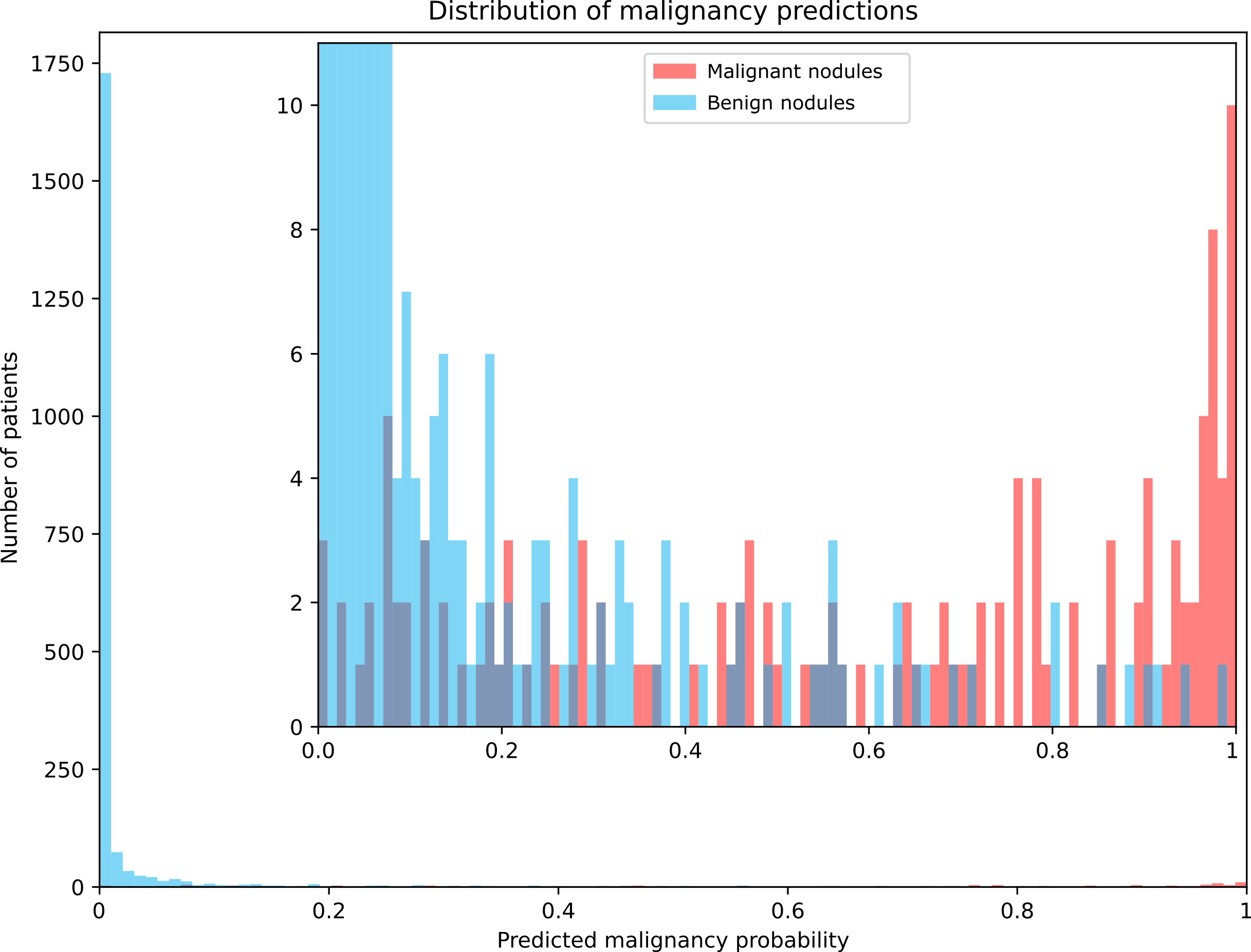}
	\centering
	\caption{\textbf{Distribution of model's patient-level Likelihood Of Malignancy on Test1}. Thanks to the calibration\cite{kuppers_multivariate_2020} of the output prediction on the training set, these output distributions optimally reflect the model's empirical accuracy.} \label{fig_sup1}
\end{figure}

\begin{figure}[!htb]
	\includegraphics[scale=1]{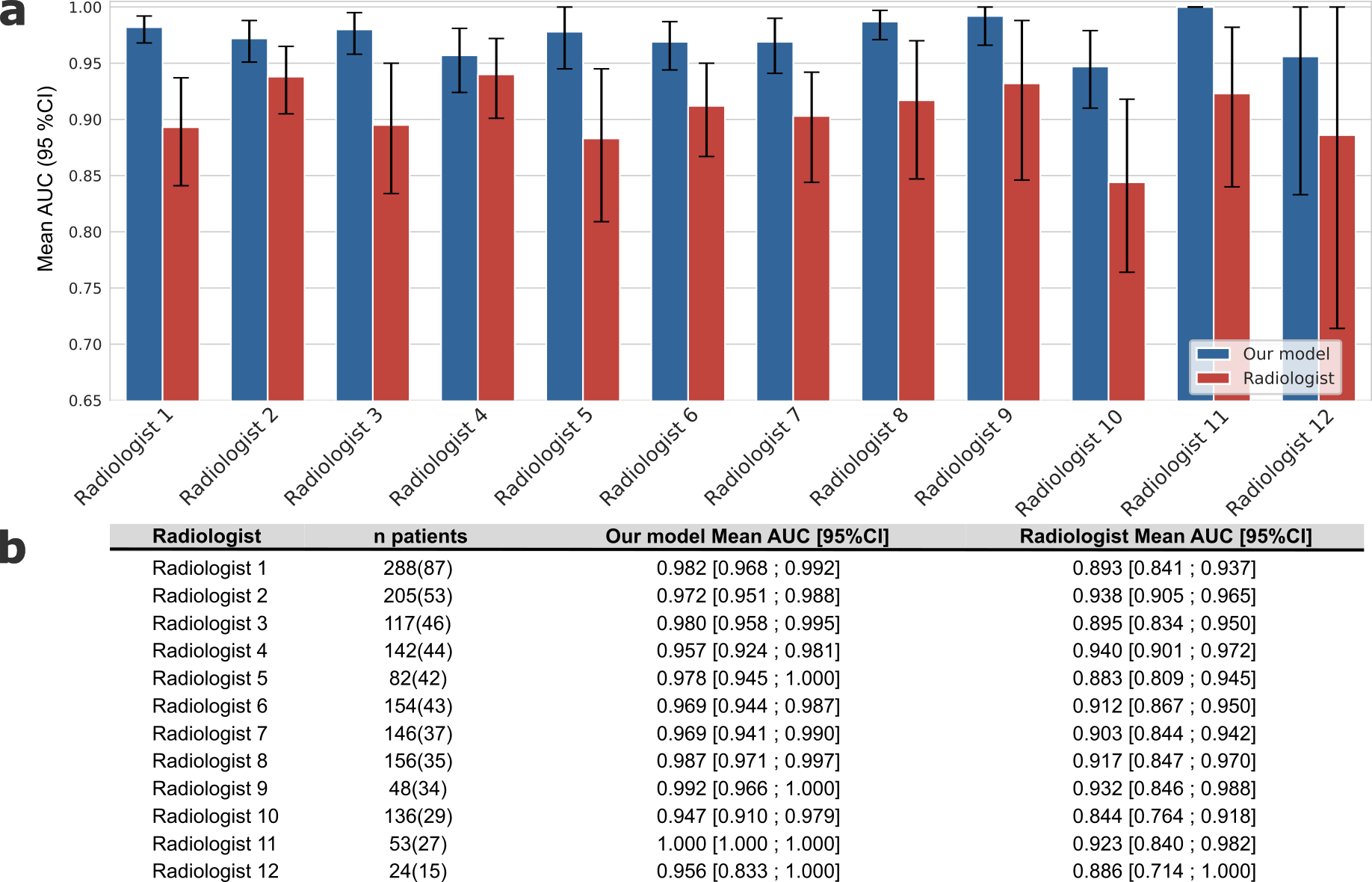}
	\centering
	\caption{\textbf{Comparison of mean AUC between our model and individual radiologists:} \textbf{a.} Mean AUCs over 5,000 bootstraps of our model and of each of the 12 radiologists on the same scan subsets of Test3, as shown in  Fig.3. Vertical bars indicates the 95\% CI. \textbf{b.} Values of mean AUCs and 95$\%$ CI illustrated in \textbf{a}. The sample size (n: number of patients) for each annotator with the number of cancer cases indicated in parenthesis. The Welch t-tests on AUC distributions over the 5,000 bootstraps confirm our model's non-inferiority of our model AUC over radiologists in all cases, with $p<0.0001$.} \label{fig_sup2}
\end{figure}


\begin{figure}[h!]
	\includegraphics[scale=0.8]{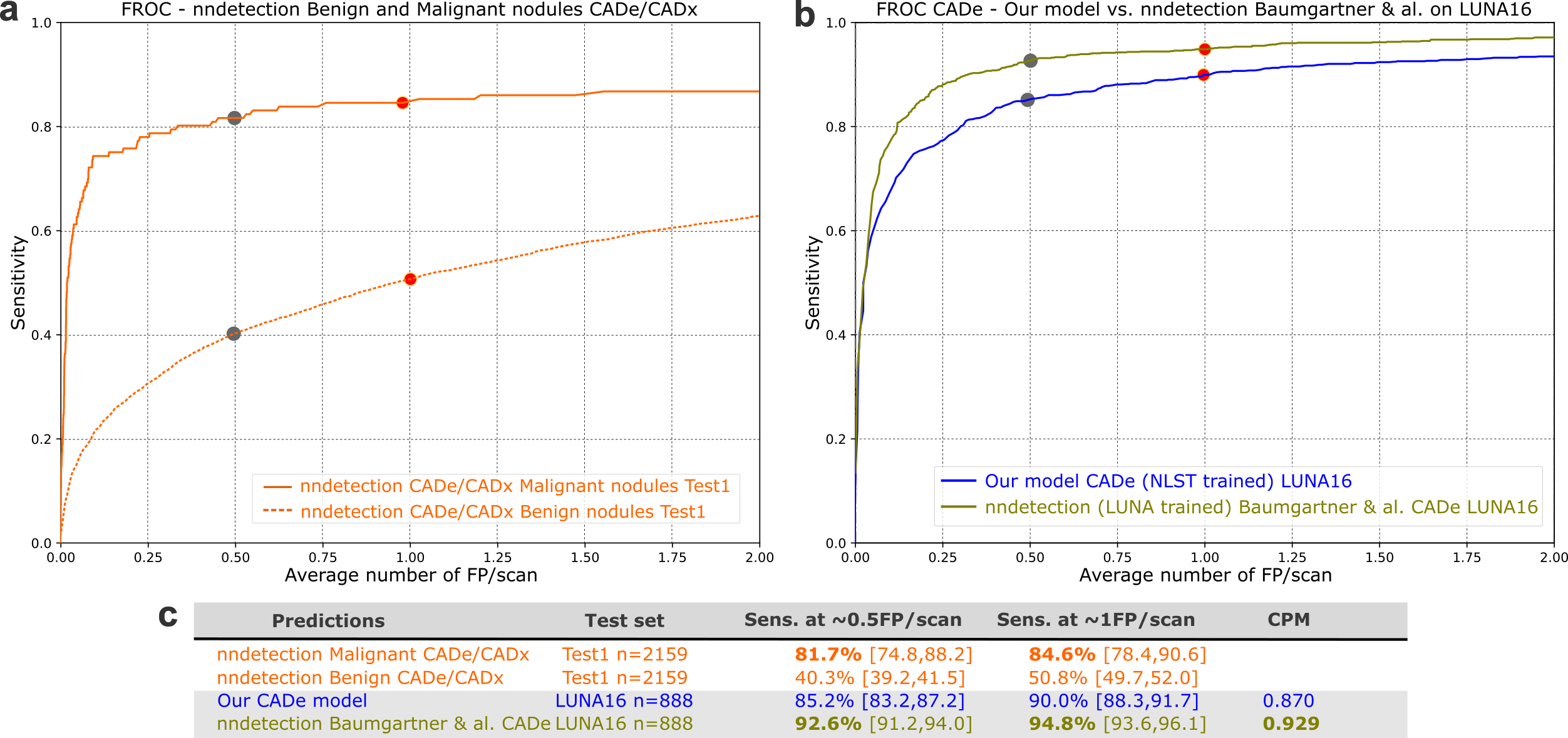}
	\centering
	\caption{\textbf{Comparison of detection performances: benign vs. malignant nodules (nnDetection), \& performance comparison of our model vs. original nnDetection on LUNA16 challenge:} \textbf{a.} The FROC curves of the nnDetection CADe/CADx model retrained on NLST (Train3) for Benign and Malignant nodules detection tasks as separate classes on Test1 (the Malignant nodule detection curve is the same as in Fig.4\textbf{a}). \textbf{b.} FROC curve of our CADe model trained on NLST alongside the nnDetection model trained by Baumgartner \textit{et al.}\cite{baumgartner_nndetection_2021} on LUNA16, on the LUNA16 nodule detection challenge dataset. The Competition Performance Metric (CPM) of 0.929 achieved in this study for the Baumgartner \textit{et al.} model reproduces their published CPM of 0.930, thereby "outperforming all previous methods on the nodule-candidate-detection task"\cite{baumgartner_nndetection_2021}. \textbf{c.} Table of the mean sensitivity over 5,000 bootstraps samples (in percent) at FROC OPs closest to 1 and 0.5 mean FP/scan for the various predictions in \textbf{a},\textbf{b}, with 95\% CI over 5,000 bootstraps samples. The CPM for LUNA16 challenge task are also included.} \label{fig_sup3}
\end{figure}

\begin{figure}[!htb]
	\includegraphics[scale=0.9]{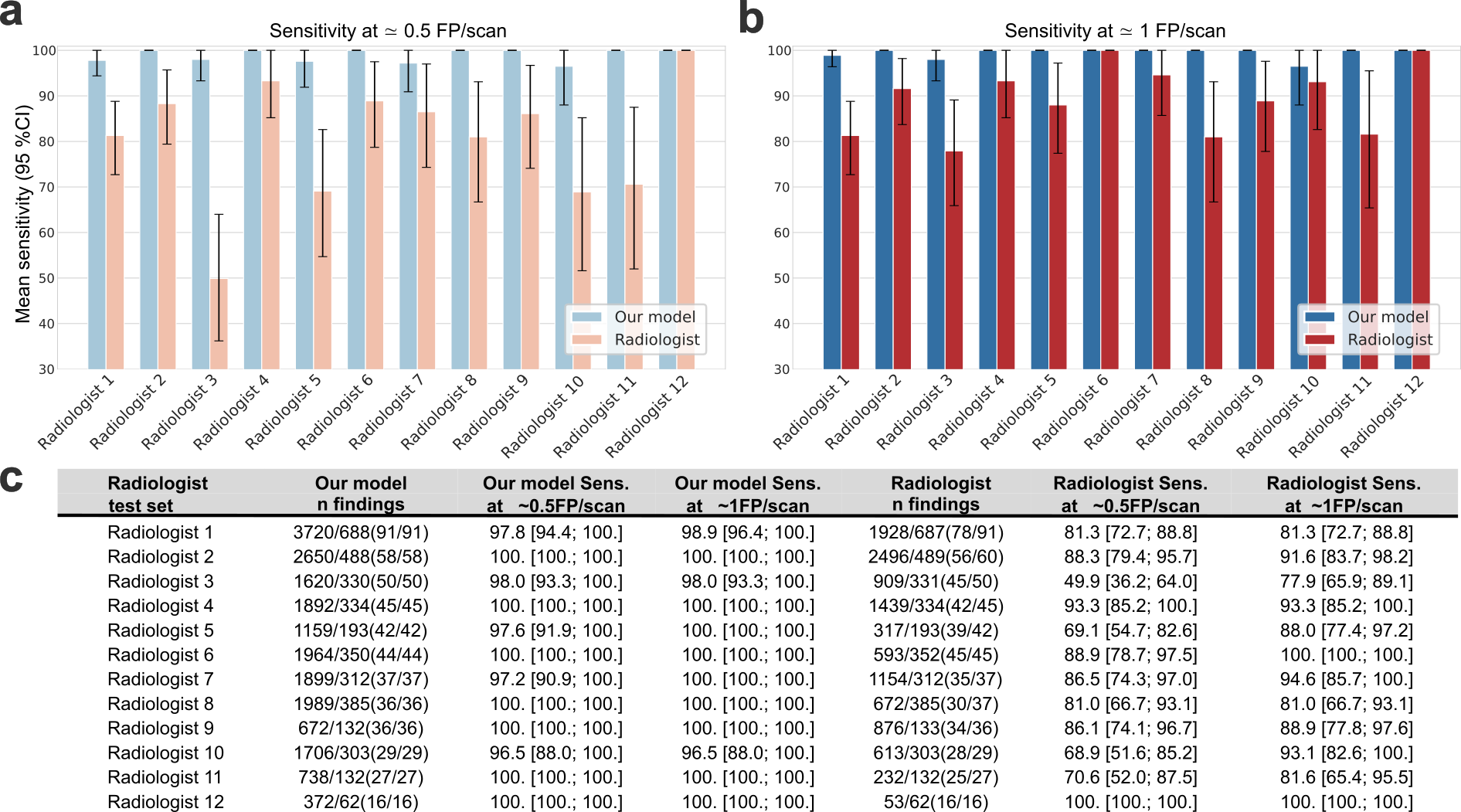}
	\centering
	\caption{\textbf{Detection Sensitivity comparison at fixed FP/scan: our model vs. individual radiologists} Mean sensitivity over 5,000 bootstraps at the closest OP to $\approx$ 0.5 in \textbf{a.} and at the closest OP to $\approx$ 1 FP/scan in \textbf{b.}, for our model and the 12 radiologists on the same subsets of Test3 corresponding to the FROCs shown in Fig.5. Vertical bars indicate the 95\% CI. \textbf{c.} Values of mean sensitivity and 95$\%$ CI illustrated in a. and b. The sample size (n: number of findings: TP+FP of CADe) and number of findings in GT (CADe TP+FN of CADe) for each annotator and the model are provided, with malignant finding cases indicated in parenthesis. The Welch t-tests on sensitivity distributions demonstrate the non-inferiority of the sensitivity of our model over each radiologist in all cases, with $p<0.0001$ both at 0.5 and 1FP/scan—except for Radiologist 12 whose small sample size leads to saturation in this range. Notably, at lower FP/scan our model still outperforms Radiologist 12, as shown in Fig.5. Radiologist's count of finding in GT can slightly exceed that of our model in GT (which matches the number of nodules in GT for malignant nodules), since radiologists may detect multiple findings with different scores for one same nodule in GT. Additionally, the closest OPs to 0.5 and 1 FP/scan on the radiologists' curves can sometimes deviate far from 0.5 and 1 FP/scan value, as shown in Supplementary Fig.5, due to malignancy scores being discretized on a 1–10 scale, limiting direct OP comparisons. Finally, the number of findings retrieved by our model, as presented in the table, may be excessively large and arbitrary, and should be restricted in practice by setting a threshold to a low FP/scan threshold (e.g., 2 FP/scan or below) without compromising performance, especially in the asymptotic constant regime.}  \label{fig_sup4}
\end{figure}

\begin{figure}[!htb]
	\includegraphics[scale=0.5]{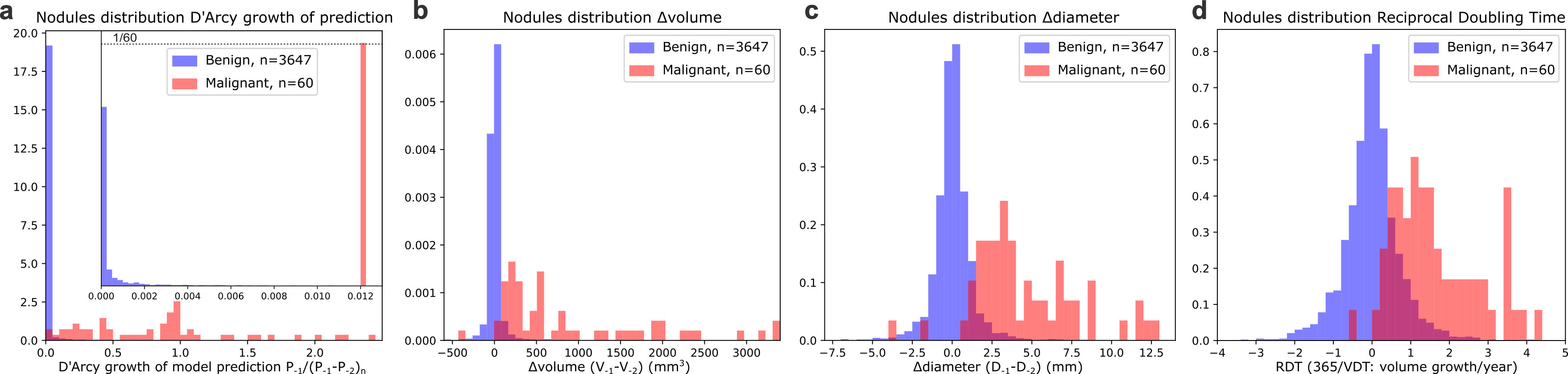}
	\centering
	\caption{\textbf{Distributions of Malignant and Benign probability corresponding to the ROCs in Fig.6a:} of D'Arcy Thompson growth of our model prediction with a zoom on small values (\textbf{a}), of $\Delta$volume (\textbf{b}), of $\Delta$diameter (\textbf{c}) and of RDT (\textbf{d}). For clarity, Benign and Malignant distributions are independently normalized to an integral of 1, and for $\Delta$diameter distributions, two nodules with outlying large delta were excluded.}\label{fig_sup5}
\end{figure}  

\begin{figure}[!htb]
	\includegraphics[scale=0.8]{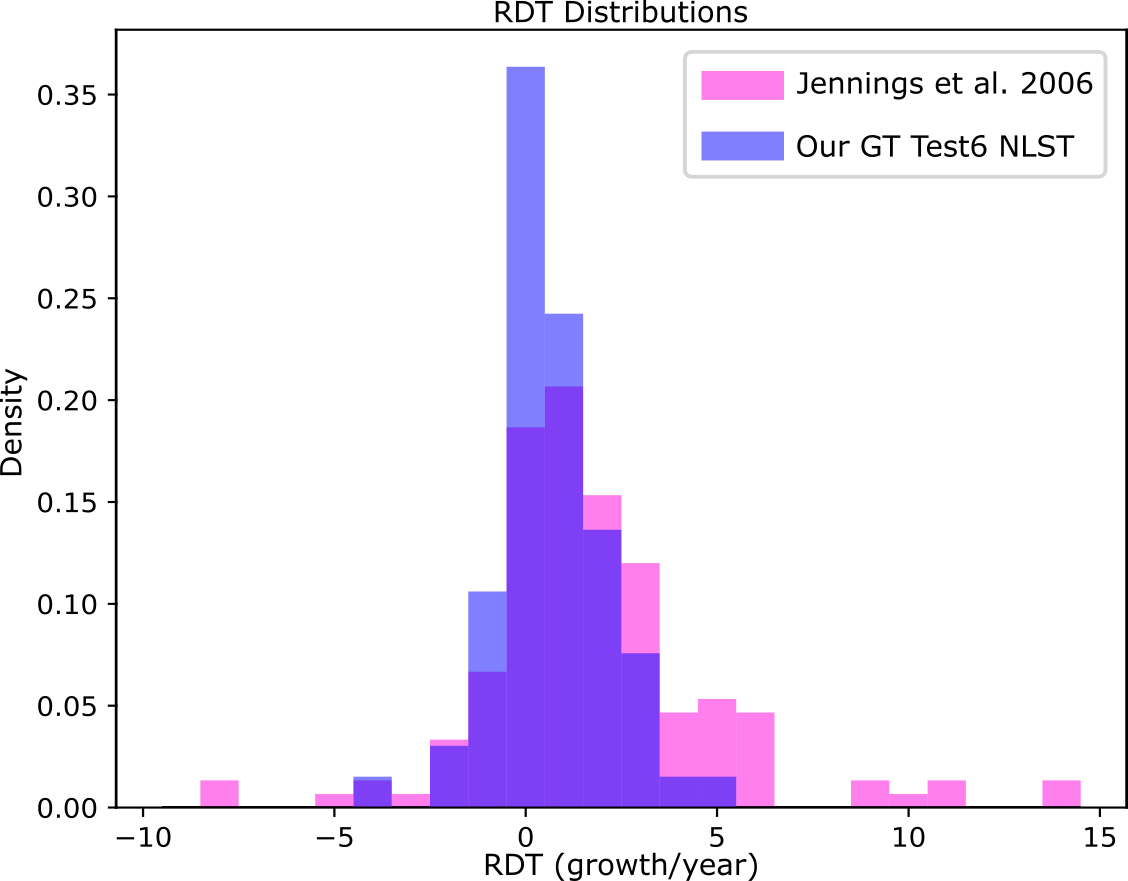}
	\centering
	\caption{\textbf{Comparison of the distribution of cancer patient RDT of Jennings \textit{et al.}\cite{jennings_distribution_2006} with the RDT of our GT over the nodules longitudinally paired in GT of Test6} (redrawn from\cite{jennings_distribution_2006}). }\label{fig_sup6}
\end{figure}

\begin{figure}[!htb]
	\includegraphics[scale=0.8]{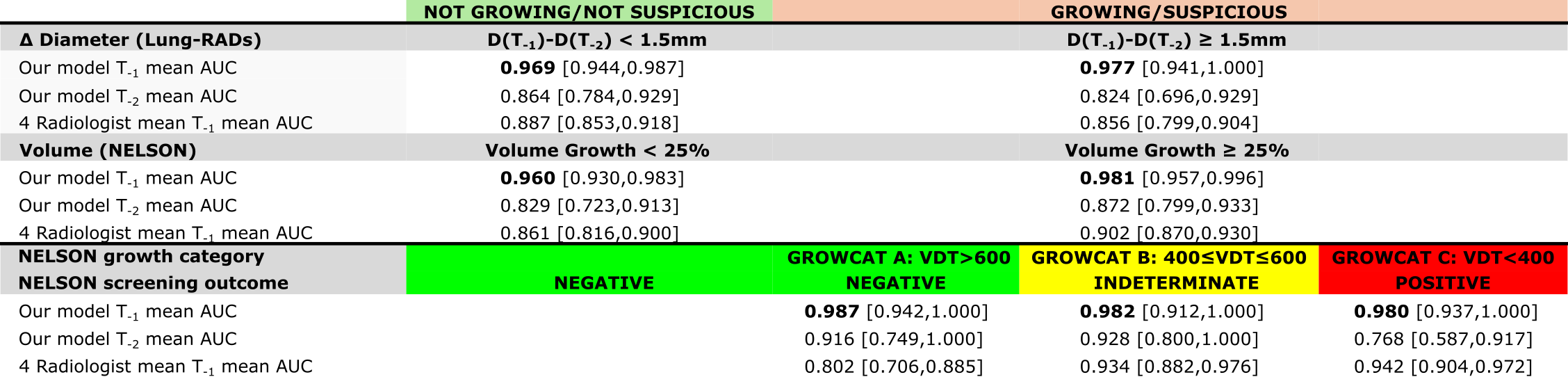}
	\centering
	\caption{Mean ROC-AUC and 95\% CI over 5,000 bootstraps samples for prediction by our model at $T_{-1}$, at $T_{-2}$ and the mean of 4-Radiologists at $T_{-1}$ on the same intersecting subgroup of Test6 and Test3, for the various subgroups of growing nodules category, $\Delta$diameter below or above 1.5 mm (Lung-RADS\textsuperscript{\textregistered}), volume growth below or above 25\% (NELSON\cite{xu_nodule_2006}) and with VDT above 600, between 400 and 600 or below 400 days, corresponding to the NELSON's GROWCAT A,B and C categories, respectively\cite{xu_nodule_2006}. }\label{fig_sup7}
\end{figure}  

\begin{figure}[!htb]
	\includegraphics[scale=0.8]{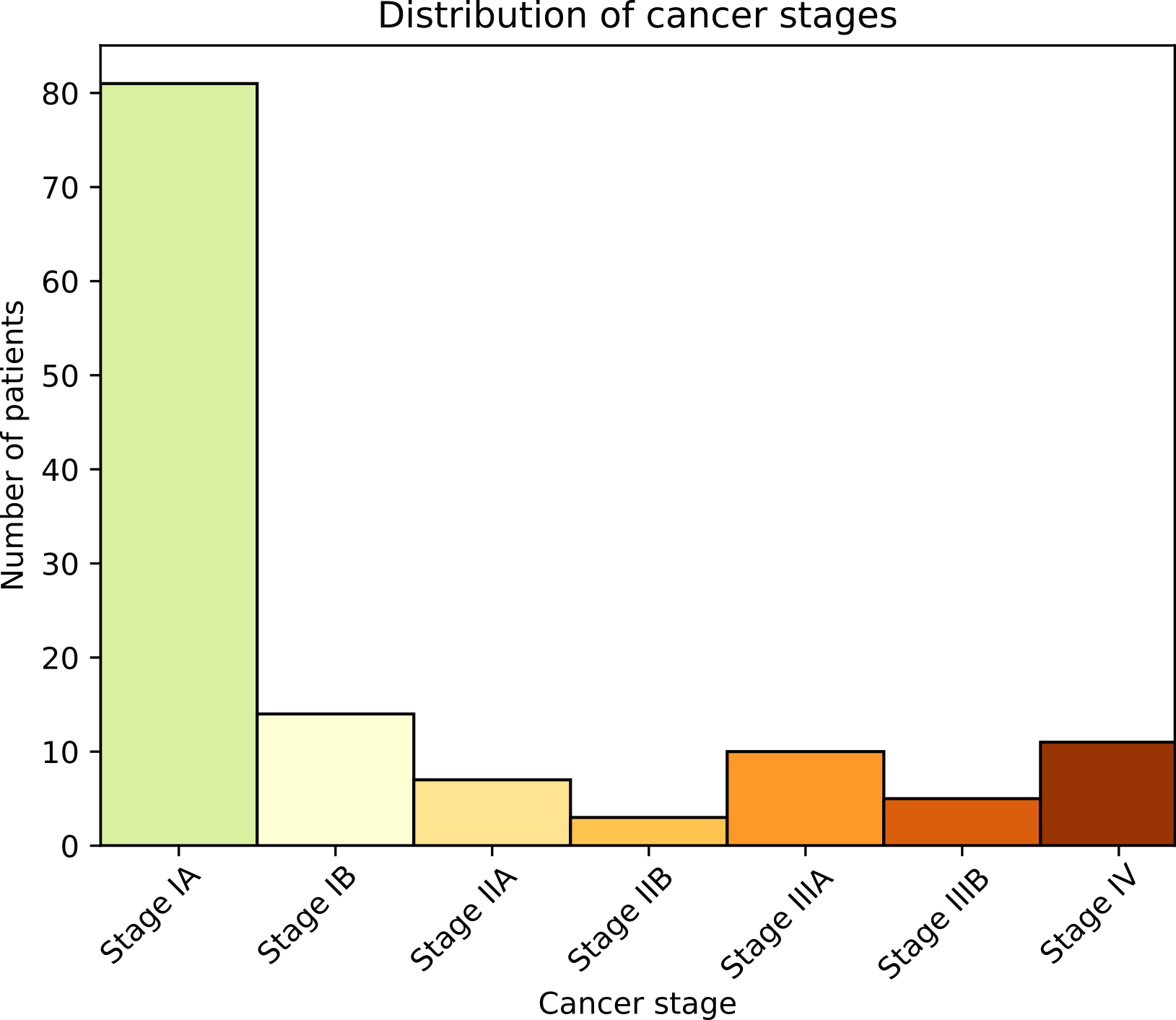}
	\centering
	\caption{The distribution of cancer stages across patients in Test1.} \label{fig_sup8}
\end{figure}

\section{Supplementary methods} \label{sup_met}

\subsection{Data}\label{sup_met_data}

\subsubsection{Annotations and GT}  

\paragraph{NLST GT annotation campaign:} 
The annotation campaign was conducted by Quantified Imaging, a medical consulting and annotation company. The annotations were conducted by one expert medical doctor specialized in chest radiology and one expert radiologist, both of whom followed a predefined annotation protocol (summarized below).    
The clinician annotations process consisted of two main tasks:  
\begin{itemize}
 \item \textbf{Task of nodule localization, segmentation and longitudinal tracking:} This task consists in detecting and semi-manually segmenting the solid component of all segmentable nodules that can be identified in the scan. It also includes linking longitudinally identical nodules across multiple TPs. Radiologists completed this task with LMS software (FDA approved, commonly used for VDT quantification\cite{becker_randomized_2015}), that provides a semi-automatic "one-click" segmentation of nodule solid components, originally designed for RECIST quantification. When a radiologist disagreed with the semi-automatic segmentation, they were required to adjust the segmentation manually.
 \item \textbf{Disambiguation task:} This task consists in linking each annotated nodule to its corresponding entry in the NLST "Spiral CT Abnormalities dataset". Each nodule not referenced in NLST records was annotated with a new GT label. For cancer patients at the time of diagnosis, malignant nodules were localized according to biopsy-confirmed outcome (size, grade, position) and other characteristics (z position, lobe position) provided by NLST. When multiple highly suspicious nodules were found at the cancer location indicated by NLST, or when no site was specified but a highly suspicious nodule was present elsewhere, the radiologist also classified the nodule as malignant.
\end{itemize}
As a result of this annotation protocol, an augmented pool of nodules is localized in each scan, associated with a mask, tracked across TPs under a unique ID, and linked to their NLST GT (or assigned a new GT if not previously recorded in NLST). Among the GT entries for each abnormality in NLST (notably in the "Abnormalities" dictionary in NLST), the 'Abnormality description' field (\textit{sct ab desc}) specifies lesion types, which can be classified into clinically distinct abnormality categories: 
\label{nlst_types}
\begin{itemize}
    \item Parenchymal lung nodular lesions: 51="Non-calcified nodule or mass (opacity >= 4 mm diameter)"; 52="Non-calcified micronodule(s) (opacity < 4 mm diameter)"; 53="Benign lung nodule(s) (benign calcification)"; 62="6 or more nodules, not suspicious for cancer (opacity >= 4 mm)";  
    \item Parenchymal lung lesion non-solid or with sometimes pseudo-nodular appearance: 54="Atelectasis, segmental or greater"; 58="Consolidation"; 61="Reticular/reticulonodular opacities, honeycombing, fibrosis, scar";   
    \item Parenchymal Lung lesion without nodular shape: 59="Emphysema"; 55="Pleural thickening or effusion";        
    \item Non parenchymal lesions: 56="Non-calcified hilar/mediastinal adenopathy or mass (>= 10 mm on short axis)"; 57="Chest wall abnormality (bone destruction, metastasis, etc.)"; 60="Significant cardiovascular abnormality"; 63="Other potentially significant abnormality above the diaphragm"; 64="Other potentially significant abnormality below the diaphragm"; 65="Other minor abnormality noted". 
\end{itemize}   
For annotation convenience, all NLST abnormality types 52 and 62, as well as newly-discovered nodular parenchymal lesions were pooled in a single class, labeled 50.

\begin{figure}[h!]
	\includegraphics[scale=0.9]{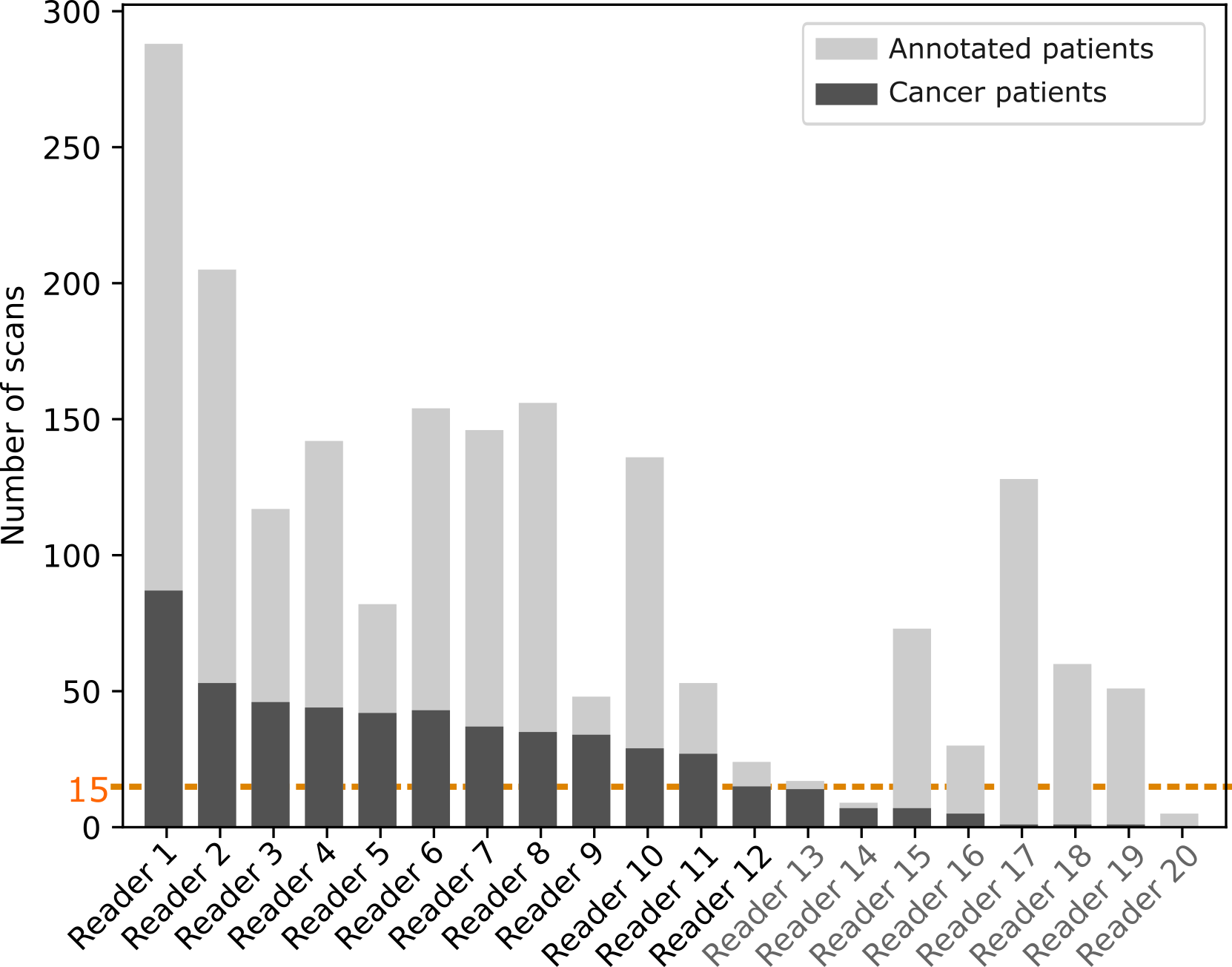}
	\centering
	\caption{\textbf{Number of annotated scans per reader:} Readers are ranked by number of annotated cancer cases. The 12 readers who annotated at least 15 scans with cancer and retained for the performance comparison to single reader, are labeled in black; all others are labeled in grey.} \label{fig_sup9}
\end{figure}

\paragraph{NLST LIDC-like radiologist assessment annotation campaign:} 
In a second step, we conducted a radiologist annotation campaign covering Test3—a subset of Test1—to evaluate performance on a malignant nodule detection task equivalent to a CADe/CADx assessment. Independent readers—practicing radiologists with a valid medical license and at least one year of experience interpreting chest CT scans—were recruited through an independent company, Ingedata, which managed the annotation campaign. To optimize flexibility, cost and time of the annotation campaign, we adapted the k-fold Split-Plot Multi-Reader Multi-Case (MRMC) protocol \cite{obuchowski_multi-reader_2012}. The new protocol relaxes the constraint that each reader assess an identical number of cases, thereby accommodating individual availability, while ensuring each case/scan is annotated independently exactly four times, by different readers. Although statistically less robust than classical MRMC design, this protocol is sufficient for our current objectives and claims; a comprehensive MRMC study is ongoing and will be published separately. Notably, as stated in the Methods ROC section, the mean ROC and AUC calculations across radiologists assume radiologist equivalence and permutation invariance. As a result, Test3 was annotated by 20 different readers, each assessing between five and 297 patients. The distribution of scan-patient counts per annotator is provided in Fig.\ref{fig_sup9}. This dataset (named LIDC-annot-NLST501) is made publicly available with the publication of this paper.

\begin{figure}[h!]
	\includegraphics[scale=0.9]{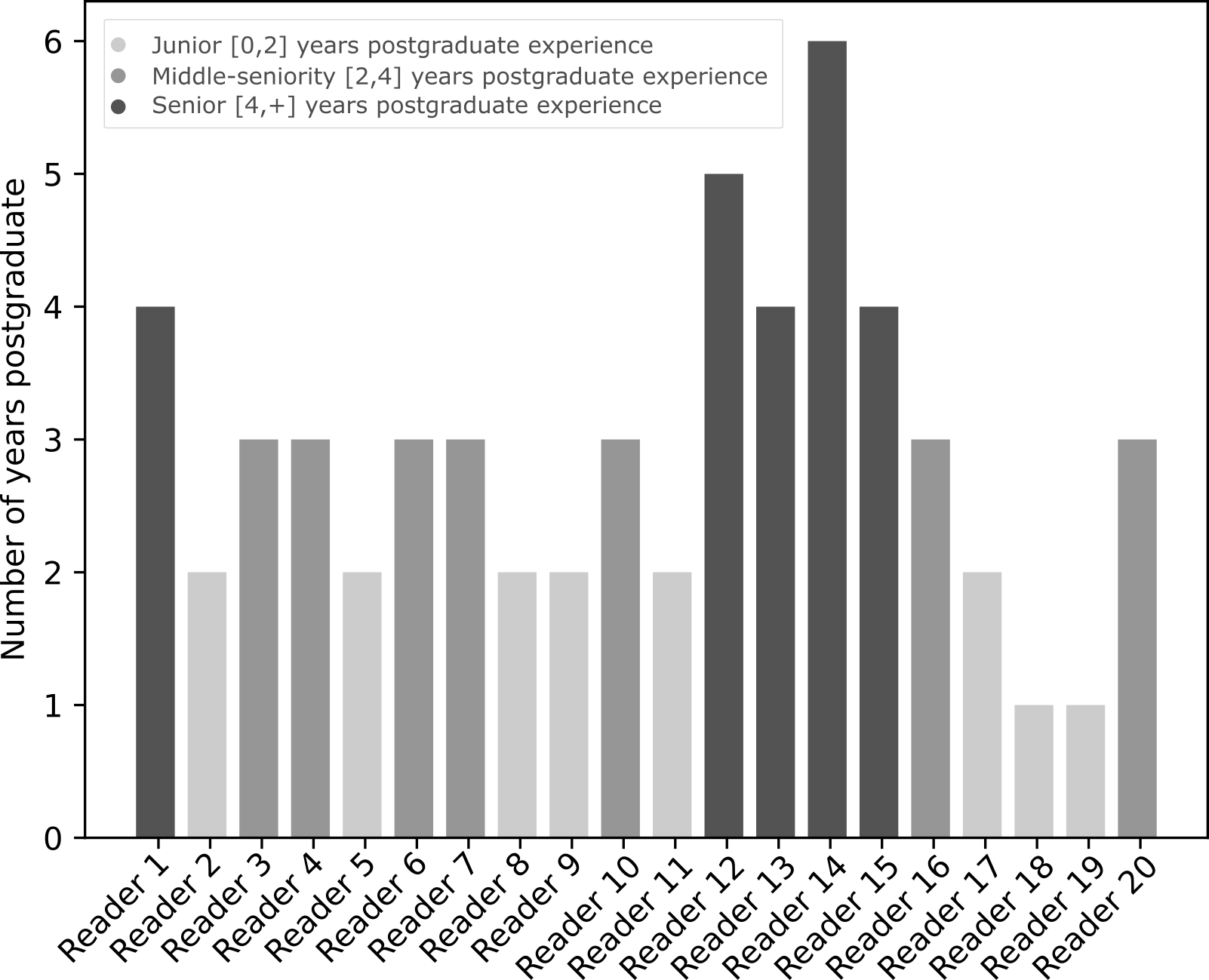}
	\centering
	\caption{\textbf{Reader experience and seniority:} Number of postgraduate years of radiologist experience per reader, categorized into Junior, Middle-Seniority, and Senior.} \label{fig_sup10}
\end{figure}

Following the seniority classification proposed in\cite{he_can_2021}—which defines junior radiologists as those with $\leq2$ years postgraduate experience, mid-level seniors with 2–4 years, and seniors with $\geq4$ years— the 20 radiologists were distributed into the following cohorts: 8 juniors; 7 mid-level seniors; and 5 seniors. This distribution of postgraduate experience across the 20 readers, illustrated in Fig.\ref{fig_sup10}, exhibited no significant correlation with the number of scans annotated per reader (correlation coefficient $\rho=-0.208$, p=0.38). 

The annotation protocol was designed to align with and adhere to the LIDC-IDRI annotation protocol and methods, as the LIDC-IDRI dataset has become a benchmark in the field and its protocol was developed by a panel of expert radiologists\cite{armato_lung_2011-1,the_tcia_team_annotation_2020}. Radiologists were blinded to the NLST clinical data and instructed to detect nodules and provide a subjective assessment of the likelihood of malignancy for each nodule, assuming the scan originated from a 60-year-old male smoker. Malignancy score ranges from one to ten, with the following definition: 
\begin{itemize}
    \item 1: Definitely benign nodules and Benign Appearance or Behavior Nodules with a very low likelihood of becoming a clinically active cancer due to size or lack of growth;  
    \item 2 to 3: Probably benign finding(s) ‐ short term follow up suggested; includes nodules with a low likelihood of becoming a clinically active cancer;
    \item 4 to 5: Suspicious Findings for which additional diagnostic testing is recommended; 
    \item 6 to 10: Very Suspicious Findings for which additional diagnostic testing and/or tissue sampling is recommended or necessary; 
\end{itemize}
The assessment from 2 to 3, 4 to 5, and from 6 to 10 is purely subjective, representing the perceived degree of suspicion based on the visual features observed.

\paragraph{Lung-RADS\textsuperscript{\textregistered} radiologist assessment and Ardila \textit{et al.} model prediction:}
We use the data provided by Ardila \textit{et al.}\cite{ardila_end--end_2019} in their supplementary material, that comprises their model's predictions alongside Lung-RADS\textsuperscript{\textregistered} 2019 assessments of lesions by six radiologists, both with and without prior TP information (see\cite{ardila_end--end_2019} for the methodology). Comparisons with our model's predictions was carried out at the patient level on the same patient set. To further enhance comparability, in the minority of cases where Ardila \textit{et al.} did not use the same TP (for non-cancer patients) or series (kernel) as our selection, we substituted with an alternate kernel from the same TP where available. If not, we used the scan from the closest available TP. 

\paragraph{Independent cohort GT annotation campaign:} The IC was assembled from three data sources: images, paired with semi-structured clinical data collected by the AIR study (EU)\cite{leroy_circulating_2017-1}, in conjunction with images collected by two US data providers. To ensure a high degree of precision, the annotation protocol implemented a consensus truthing process based on three readers, and proceeded in three stages. However, this protocol could not be strictly identical to the NLST protocol due to discrepancies in available clinical information. Initially, two concurrent readers performed localization and measurement of all nodules and abnormalities, identifying whether or not these constituted lesions of interest (i.e., within the $[4,40]$mm range). When present, each reader was required to specify solidity and margins. For cancer patients, readers were required to classify the nodule as a candidate or confirmed cancer (or rule out that possibility) based on the available clinical data. For each patient, readers were to deliver no more than 15 lesions. A second stage addressed discrepancies among the initial concurrent readings. Following lesion pairing, an independent reader (hereafter adjudicator) reviewed each annotation conflict, confirming or rejecting nodules by location. Upon accepting a single-read nodule, the adjudicator also recorded their own assessment of lesion characteristics. A separate adjudicator performed an independent review of the lesion characteristics alone. As a result nodule location and characteristics were established by consensus of at least two of the three radiologists (2/3). In the final stage, all cancer-confirmed or candidate nodules, along with a representative sampling of benign nodules, underwent concurrent segmentation by the three readers. The GT segmentation was defined by a two-thirds consensus across the three segmentation masks. All readers were recruited and overseen by Ingedata, and were current practicing radiologists with a minimum of two years’ experience interpreting chest CT. Adjudicators were selected on the basis of seniority.\\ 

\subsubsection{Inclusion/exclusion}\label{sup_met_inclu_exclu} 

\paragraph{NLST inclusion/exclusion}:\\
\textbf{Exclusion of unavailable CT scans:} The CT arm of NLST, excluding all X-Ray images, comprised 26,722 patients. Of these, only 12,810 patients—including 1,047 cancer patients and 11,763 non-cancer patients with nodules—were eligible for download without failure.\\

\textbf{Kernel selection: } For all those patients, for each of the 3 TP scans ($T_{-1}$, $T_{-2}$, $T_{-3}$)—when available—only a single CT scan was selected among the multiple scans acquired for each TP with varying kernels and acquisition parameters (up to ten scans per patients per TP). The selection criteria for this single scan closely follows those established by Ardila \textit{et al.}\cite{ardila_end--end_2019}: prioritizing Hard Kernels (best signal-to-noise ratio for parenchymal nodule detection and characterization); eliminating localizers; excluding scans with insufficient resolution or number of images. The exact inclusion/exclusion rules for kernel selection are as follows\cite{ardila_end--end_2019}:  
(1) Exclude all localizers; (2) For each case, select the highest ranking kernel according to the following lists (no cases involve more than one manufacturer ):
\begin{itemize}
\item Siemens: (1) B50f; (2) B45f; (3) B50s; (4) B40f; (5) B41s; (6) B60f; (7) B60s; (8) B70f; (9) B36f; (10) B35f; (11) B30f; (12) B31s 
\item GE:  (1) LUNG; (2) BONE; (3) BODY FILTER/BONE; (4) STANDARD; (5) BODY FILTER/STANDARD; (6) SOFT; (7) EXPERIMENTAL7; (8) BODY FILTER/EXPERIMENTAL7 
\item Philips: (1) D; (2) C; (3) B; (4) A  
\item Toshiba: (1) FC51; (2) FC50; (3) FC52; (4) FC53; (5) FC30; (6) FC11; (7) FC10; (8) FC82; (9) FL04; (10) FC02; (11) FC01; (12) FL01 
\end{itemize} 

\textbf{Scan Quality Control (QC) exclusion:} Scan QC implements the following inclusion/exclusions:
\begin{itemize}
\item \textbf{Scan slice thickness:} scans with slice thickness greater or equal to 5 mm were excluded from the study. 
\item \textbf{Sufficient lung coverage:} whenever a scan did not have sufficient number of images, indicating incomplete coverage of the lung, we replaced it by the next best available based on kernel ranking in the previous list and scans without suitable replacement were excluded. The QC exclusion was performed manually on the set of scans suspected of covering less than 25 cm of the body (assuming lung size should be greater than 25 cm). 
\item \textbf{Prone position:} CT scans acquired in the prone position were excluded
\item \textbf{Artifacts:} scans with artifacts in the lung altering the image quality were excluded.
\item \textbf{Incorrect scan reconstruction:} scans with missing slices were excluded.
\end{itemize}

\textbf{Paremchymal nodule and truthing inclusion/exclusion:} As a result, each patient contributed a single selected, readable scan for each TP. In total $9,183$ non-cancer cases and $695$ cancer cases were submitted for annotation to Quantified Imaging (the contracted annotation company). 
For Train3 (CADx training set) and Test1, we included one additional exclusion criterion following the annotation, to retain only correctly localized and diagnosed malignant parenchymal nodules.
Among cancer patients, 600 were identified as having at least one disambiguated malignant lesion: for 95 patients, it proved infeasible to detect the cancerous lesion on the scan. This was due to a number of factors, including: discrepancy between clinical diagnostic information and CT scan data (e.g., size or location); the time lapsed between the invasive diagnostic procedure confirming the cancer and the date of scan acquisition sometimes exceeded one year; and additional image quality issues (e.g., motion artifact) that could render the CT scan reading inconclusive.\\
For nodule definition, we adhered to the guidelines of the Fleischner society\cite{bankier_fleischner_2024}. Our study focuses exclusively on parenchymal nodules, so it includes/excludes the following:
\begin{itemize}
    \item Solid and part-solid inclusion: patients with solid and part-solid malignant nodules were included; patients with pure GGO malignant lesions were excluded.
    \item Parenchymal inclusion: patients with parenchymal malignant nodules were included; patients with mediastinal or hilar masses/malignant lesions were excluded.
    \item Diameter range inclusion $[4,30]$mm: for the tests sets, patients with malignant nodules in the range of $[4,30]$mm diameter on the segmentation GT were included, all patients with malignancies outside this range were excluded. For the training sets, a broader inclusion range of $[3,40]$mm diameter was considered , in order to prevent performance degradation and enhance generalizability.
\end{itemize}
In the original lesion classes ontology of NLST introduced in previous section, this inclusion corresponds to a lesion selection of type (\textit{sct ab desc}) 51, 53, 62, and 50.
As a result of these inclusion/exclusion criteria based on intended use, and the union of the Train3 and Test1, we obtained a cohort of 448 cancer patients and 9,183 non-cancer patients.\\ 
For Train1 (the CADe training set) only patients with purely ground glass nodules were excluded, resulting in 543 cancer patients and 7,156 non-cancer patients.

\textbf{TP selection:} for Train1 (CADe) all available TPs were selected. For Train3 and all tests sets, among all available TPs for each cancer patient, we select the scan acquired at time of diagnosis, if available, or otherwise the nearest scan acquired within one year prior to the diagnosis confirmation. For non-cancer patients, the earliest available TP was selected to maximize follow-up duration and ensure cancer-free status, while also increasing the sample size of non-cancer in LCS programmes due to the high dropout rates frequently observed during such programmes.\\ 
For longitudinal analyses and for similar reasons, for non-cancer patient, only the first two consecutive TPs available (approximately one year apart) were selected. For cancer patients, only those with scans acquired within $[0,12]$ months prior to time of diagnosis ($T_{-1}$), and $[12,24]$ months prior to time of diagnosis ($T_{-2}$) were included. Accordingly, patients diagnosed with cancer during the first year of the LCS programme.  

\paragraph{LIDC inclusion/exclusion}:\\ 
In line with previous studies, we only considered nodules annotated by at least four radiologists with a diameter greater than or equal to 3 mm.\\
For Train3 (CADx training): LIDC-IDRI also provides some additional clinical information for a subset of patients, including biopsy and resection results for some cancer cases, and follow-up data for some non-cancer cases. Using these data, we were able to isolate 77 cancer cases with biopsy/resection histopathological confirmation, and 36 non-cancer cases, i.e., consistent with our NLST inclusion protocol. These confirmed patients were incorporated into Train3.\\ 
For Train2 (CADx pretraining set): For each nodule, a radiologist rated malignancy on a scale from 1 to 5. We included all patients and lesions (non GG0, $[3,40]$mm mean diameter range) that were not included in Train3 (i.e., not confirmed by biopsy). The final malignancy labels were obtained by taking the median value of all radiologists' ratings. Nodules with a median of 3 were excluded, as they could not be reliably classified as benign or malignant. This selection results in a total of 656 nodules, of which 275 were labelled as malignant (54\%), and 268 as benign (46\%) with weak annotation GT.\\
For Train1 (CADe training ): we included all patients and lesions (non GG0, $[3,40]$mm mean diameter range) applying the same method and rules as Train2.  

\paragraph{IC inclusion/exclusion}:\\ 
The IC population was compiled from three sources. The first consisted of 133 patients (18 cancer patients) enrolled in the French AIR\cite{leroy_circulating_2017-1} that investigates the predictive power of blood biomarkers for lung cancer in a LCS population with COPD, following the clinical inclusion criteria provided. The second dataset included 276 US patients (92 cancer cases) purchased from the data provider SEGMED. 412 patients (199 cancer patients), also from the US, acquired via Gradient; of these, 66 patients (36 cancer cases) were reserved for a separate study. The third comprised 412 patients (199 cancer cases), also from the US, acquired via the data provider Gradient; of these 66 patients (36 cancer cases) were reserved for a separate study. Each subset underwent a two-stage screening: first, verification for compliance with our clinical protocol; second, quality control of imaging data. Following this, scans were annotated and disambiguated. The inclusion criteria used were the following:\\
\begin{itemize}
    \item aged 50 to 80 years;
    \item current or former smoker ($\geq 20$ pack-years).
    \item underwent patient screening and surveillance as part of a LCS programme; adhering to LCS guidelines.
    \item underwent LDCT screening following classification as high-risk for lung cancer.
\end{itemize}
Benign status was verified as follows:
\begin{itemize}
    \item Evidence of nodule stability for at least 12 months, documented either by imaging (follow-up scans showing no progression or regression), and/or radiological reports, explicitly stating stability or reduction in size.
    \item Negative tissue sampling within 12 months of LDCT screening scan.
\end{itemize}
Cancer status was verified as follows:
\begin{itemize}
    \item with proof of cancer within 12 months of a LDCT screening exam. Either from extraction of histopathology reports, or from mention of primary lung cancer in radiological reports (or corresponding codes from International Classification of Diseases for lung neoplasm—
 excluding carina, hilum, mediastinum, and bronchus cancers, or specific to lung cancer cell types).
    \item additional information eligible to derive cancer diagnosis, such as mention of diagnostic procedures (Cytopathology, Biopsy, etc.), mention of treatment (resection, excision, chemo/radiotherapy, etc.), or relevant patient aftercare follow-up.
\end{itemize}
The various steps of the data selection process, as illustrated in Fig.7 of the article are:
\textbf{Clinical QC exclusion:} the exclusion criteria are as follows: 
\begin{itemize}
    \item history of lung resection;    
    \item pacemaker or other intra-thoracic indwelling metallic medical devices that interfere with CT acquisition; 
    \item prior inclusion (patients/images) in AI model development;
    \item patients with only mediastinal cancer(s);
    \item patients with only ground glass cancer(s);
    \item patients with nodules, solid or part-solid, >30 mm (masses). 
\end{itemize}
\textbf{Scan QC exclusion:} Each selected image was manually reviewed for consistency with the corresponding electronic case report forms. Scans exhibiting any of the following were excluded: 
\begin{itemize}
    \item slice thickness >3 mm;    
    \item missing slices; 
    \item partial coverage of the lung; 
    \item artifacts;
    \item contrast injection.
\end{itemize}
Patients without any image suitable for analysis, were excluded.  
\textbf{Truthing exclusion}: cases were rejected when the radiologist responsible for establishing the GT was unable to resolve ambiguity regarding either the location or malignancy of suspicious lesions. In some instances, lesions observed in presumed benign cases appeared highly suspicious, prompting a verification process before definitive inclusion. These cases were flagged for clarification with the original data providers. If the site could not confirm the lesion's status—whether malignant or benign—the patient was excluded during truthing. 
\textbf{Intended use exclusion:} the following exclusion criteria were applied: 
\begin{itemize}
    \item patients with only mediastinal cancer(s);
    \item patients with only ground glass cancer(s);
    \item patients with nodules, solid or part-solid >30 mm (masses).
\end{itemize}
\textbf{Time Point Selection:} For cancer patients, the annotated scan (TP) is the nearest LDCT scan acquired within 12 months before the date of diagnosis. In benign cases, the annotated scan is the earliest available LDCT for LCS. Scan selection followed NLST criteria: original axial; prone position; most lung-like kernel with slice thickness closest to 1 mm; at most 5 mm; and with less than 1,000 slices.
Note that any deviation from the above criteria triggered a query to the originating site (e.g., for missing information). Any such queries (whether clinical or imaging related) remaining unresolved at the time of analysis led to the exclusion of the patient.

\subsection{Data analysis}\label{sup_met_Data_analysis}

\subsubsection{Competition Performance Metric (CPM):} The CPM averages sensitivity over a predefined list of FPs/scan thresholds. It was designed to enable consistent comparison by summarizing performance across multiple FROC OPs, and was notably used for the LUNA16 challenge\cite{setio_validation_2017}. The list of FP/scan thresholds used for LUNA16 challenge is $[1/8,1/4,1/2,1,2,4,8]$, and we use this definition in order to compare with LUNA16 challenge results. We adopted the same FP/scan thresholds used in LUNA16—[1/8, 1/4, 1/2, 1, 2, 4, 8]—for direct comparability with LUNA16 results.

\subsubsection{Subgroup analysis:}
As shown in Fig.2 of the article, the NLST and IC datasets exhibit very different slice thickness distributions, owing to NLST comprising older scans typically acquired at greater slice thickness. To account for this, we applied dataset-specific stratifications: $[0.5,1.5]$mm, $[1.5,2.3]$mm, $[2.3,3.5]$mm for NLST; and $[0.5,0.8]$mm, $[0.8,1.5]$mm, $[1.5,3]$mm for IC.\\
Due to insufficient sample size for performance assessment, particularly in the IC cohort ($n=2$), Canon and Philips were not reported in the subgroup analysis. Additionaly, 'Siemens healthineer' and 'SIEMENS' were pooled together.\\
Sharp, average, and soft kernel subgroups used in  Fig.2 of the article were defined according to manufacturers-provided specifications:
\begin{itemize}
    \item  Extra Sharp siemens kernel list: [ "['I70f'; '1']", "['I70f'; '2']", "['I70f'; '3']", "['I80s'; '1']", "['I80s'; '2']", "['I80s'; '3']", "['I80s'; '4']", "['I80s'; '5']", "B70f", "B70s", "B75f", "B75h", "B75s", "B80f", "B80s", "B90s", "I70f", "I80s", "Hr68f"] \\
    Extra Sharp GE kernel list: ["BONE", "Bone2", "BONEPLUS2", "BODY FILTER/STANDARD", "BODY FILTER / STANDARD", "BONEPLUS", "BODY FILTER/BONE", "BODY FILTER/EXPERIMENTAL"]\\
    Extra sharp toshiba list: ["FC55", "FC56", "FC59", "FC65", "FC57", "FC58", "FC80", "FC81", "FC82", "FC86"]
    \item Sharp Siemens list: ["I50f", "B50s", "['I50f'; '1']", "['I50s'; '2']", "I50s", "['I50s'; '1']", "B50f", "['I50f'; '2']", "['I50f'; '3']", "['I50s'; '3']", "['Bl54f'; '2']", "['Bl54f'; '1']", "['Br54d'; '2']", "['Bl54d'; '1']", "Bl54d", "['Bl54d'; '2']", "['Bl54d'; '3']", "Br54d", "['Br54d'; '1']", "['Bl54f'; '3']", "Bl54f", "['Br54d'; '3']", "['Bl56f'; '1']", "['Bl56f'; '2']", "['Bl56f'; '3']", "['Bl57d'; '1']", "['Bl57d'; '2']", "['Bl57d'; '3']", "['Bl57f'; '1']", "['Bl57f'; '2']", "['Bl57f'; '3']", "['Bl60f'; '1']", "['Bl60f'; '2']", "['Bl60f'; '3']", "['Bl64d'; '1']", "['Bl64d'; '2']", "['Bl64d'; '3']", "['Bl64f'; '1']", "['Bl64f'; '2']", "['Bl64f'; '3']", "['Bl64f'; '4']" "['Bl64f'; '5']", "['Br51f'; '1']", "['Br51f'; '2']", "['Br51f'; '3']", "['Br58f'; '1']", "['Br58f'; '2']", "['Br58f'; '3']", "['Br59d'; '1']", "['Br59d'; '2']", "['Br59d'; '3']", "['Br59f'; '1']", "['Br59f'; '2']", "['Br59f'; '3']", "['Br60f'; '1']", "['Br60f'; '2']", "['Br60f'; '3']", "['Br64d'; '1']", "['Br64d'; '2']", "['Br64d'; '3']", "['Br64f'; '1']", "['Br64f'; '2']", "['Br64f'; '3']", "['Br64f'; '4']", "['Br64f'; '5']", "B60f", "B60s", "B65f", "B65s", "Bl56f", "Bl57d", "Bl57f", "Bl57s", "Bl60f", "Bl64d", "Bl64f", "Br51f", "Br58f", "Br59d", "Br59f", "Br60f", "Br64d", "Br64f", "Tx60f"] \\
    Sharp GE list:  ["CHEST", "CHST", "HD Lung", "HD LUNG", "LUNG"]\\
    Sharp Toshiba list: ["FC52", "FC53 ", "FC53", "FC31", "FC50", "FC51", "FC51 ", "FC24", "FC30", "FC17", "FC14", "FC18", "FC19", "FC02", "FC08", "FC10", "FC11 ", "FC12", "FC13", "FC13-H", "FL01", "FL03", "FL04"]
    \item   Smooth Siemens kernel list: ["B45s", "B45f", "Br49d", "['Br49d'; '3']", "Br49f", "['Br49d'; '2']", "['Br49d'; '1']", "['Br49f'; '3']", "['Br49f'; '1']", "['Br49f'; '2']", "['B44d'; '1']", "['B44d'; '2']", "['B44d'; '3']", "['B44f'; '1']", "['B44f'; '2']", "['B44f'; '3']", "['Bf37f'; '1']", "['Bf37f'; '2']", "['Bf37f'; '3']", "['Br32d'; '1']", "['Br32d'; '2']", "['Br32d'; '3']", "['Br32f'; '1']", "['Br32f'; '2']", "['Br32f'; '3']", "['Br36d'; '1']", "['Br36d'; '2']", "['Br36f'; '1']", "['Br36f'; '2']", "['Br36f'; '3']", "['Br36s'; '1']", "['Br36s'; '2']", "['Br36s'; '3']", "['Br38f'; '1']", "['Br38f'; '2']", "['Br38f'; '3']", "['Br40d'; '1']", "['Br40d'; '2']", "['Br40d'; '3']", "['Br40f'; '1']", "['Br40f'; '2']", "['Br40f'; '3']", "['Br44f'; '1']", "['Br44f'; '2']", "['Br44f'; '3']", "['Bv36d'; '3']", "['Bv40d'; '1']", "['Bv40d'; '2']", "['Bv40d'; '3']", "['Bv40f'; '1']", "['Bv40f'; '2']", "['Bv40f'; '3']", "['I26f'; '1']", "['I26f'; '2']", "['I26f'; '3']", "['I30f'; '1']", "['I30f'; '2']", "['I30f'; '3']", "['I30s'; '1']", "['I30s'; '2']", "['I30s'; '3']", "['I31f'; '1']", "['I31f'; '2']", "['I31f'; '3']", "['I31s'; '1']", "['I31s'; '2']", "['I31s'; '3']", "['I40f'; '1']", "['I40f'; '2']", "['I40f'; '3']", "['I41f'; '1']", "['I41f'; '2']", "['I41f'; '3']", "['I41s'; '1']", "['I41s'; '2']", "['I41s'; '3']", "['I41s'; '4']", "['I44f'; '1']", "['I44f'; '2']", "['I44f'; '3']", "B08f", "B08s", "B10f", "B10s", "B19f", "B19s", "B20f", "B20s", "B25f", "B26f", "B29f", "B29s", "B30", "B30f", "B30s", "B31f", "B31s", "B35f", "B35s", "B36d", "B36f", "B36s", "B39f", "B39s", "B40f", "B40f", "B40s", "B41f", "B44d", "B44f", "B46f", "B46s", "Bf37f", "Br32d", "Br32f", "Br36d", "Br36f", "Br36s", "Br38f", "Br40d", "Br40f", "Br44f", "Bv40d", "Bv40f", "I26f", "I30f", "I30s", "I31f", "I31s", "I40f", "I41f", "I41s", "I44f", "T20f", "T20s", "Tr20f"]\\
    Smooth GE kernel list: ["STANDARD", "Detail2", "Detail", "Veo", "EXPERIMENTAL", "SOFT"]\\
    Smooth Toshiba kernel list: ["FC01"]
\end{itemize}

\subsection{Model}
The number of parameters for each DNN submodule is reported in Fig.8 of the article. For reproducibility\cite{haibe-kains_transparency_2020}, the main hyperparameters of all DNN submodules are listed in Table\ref{tab_sup_model_hyperparam}.\\

 \begin{table}[h!]
	\centering
	\caption{\textbf{Models hyperparameters:} List of the main hyperparameters for each DNN sub-modules. LR stands for Learning Rate, SGD stands for pytorch's mini-batch Stochastic Gradient Descent, MNO stands for Momentum and Nesterov Optimization,*Mini-batch size. }\label{tab_sup_model_hyperparam}
\begin{tabular}{ l | c | c | c | c | c | r }
  \hline			
  \rowcolor{gray!40} Module Hyperparameters & LR & LR schedule & Optimizer & Momentum & Batch size & Epochs \\
  \hline
   Lung segmentation & 1e-4 & None & Adam & Defaults & 2 & 100 \\
   Nodule detection (CADe) & 1e-2 & None & SGD & MNO & 4* & 50 \\
   Nodule segmentation (CADe) & 1e-3 & None & Adam & Defaults & 32 & 100 \\
   3D CNN (CADx) & 1e-4 & None & Adam & Defaults & 3 & 200 \\
   2D CNN (CADx) & 1e-4 & None & Adam & Defaults & 80 & 300 \\
   Full-CT scan (CADx) & 1e-5 & None & AdamW & Defaults & 8 & 300 \\
  \hline  
\end{tabular}
\end{table}

\subsubsection{Lung segmentation module} 
Our approach leverages chest CT scans which are preprocessed using a unified pipeline to
ensure data uniformity. Volumes are resampled to $128\times128\times128$ voxels using trilinear interpolation, and Hounsfield Units (HU) intensity values are clipped to $[-1024,150]$, following conventional lung windowing protocols. These values are then normalized to the $[0,1]$ range. Output masks are subsequently resampled to the original dimensions through trilinear interpolation. Our model is an adaptation of the 3D SegResNet  architecture\cite{myronenko_3d_2019-1}, originally designed for 3D brain tumor segmentation. This relatively lightweight model was chosen to accommodate a manually re-annotated subset of the LUNA16 public dataset\cite{setio_validation_2017}, which provides chest CT scans annotated with lung masks. Subset 0 is used for training and validation, while subset 1 is reserved for testing. The architecture employs residual units in an encoder-decoder design, mitigating gradient instability during training. Empirical optimization led to the selection of a 13-dimensional initial feature space.
Training hyperparameters are listed in Table~\ref{tab_sup_model_hyperparam}. We use a SoftDice loss function, derived from the Dice similarity coefficient, to capture segmentation accuracy by comparing probabilistic predictions to ground-truth annotations. Small, isolated misclassified regions are removed by retaining only the largest connected component for each predicted class. This refinement proves particularly beneficial when processing low-quality input scans, yielding substantial gains in segmentation accuracy.

\subsubsection{Nodule detection module (CADe)} 
The CADe architecture builds upon nnDetection\cite{baumgartner_nndetection_2021} that implement a RetinaNet architecture\cite{jaeger_retina_2018-2} and leverages voxel-level segmentation ability. This one-stage, anchor-based detector, is favoured for its computational efficiency. At its core, the RetinaNet model combines ResNet for deep feature extraction and feature pyramid networks (FPNs) to construct a multi-scale feature level. The layers were adapted for 3-dimensional input volumes. The network has two main output branches: detection and segmentation. In turn, the detection branch includes two smaller branches: a classifier (predicting object class inside a bounding box); and a regressor (predicting bounding box offsets for each anchor). These sub-branches are trained using Binary Cross-Entropy loss and generalized IoU loss, respectively. The segmentation branch predicts the voxel-wise volume mask and is trained with Dice and Cross-Entropy loss. Adaptive Training Sample Selection is employed to match anchors with bounding boxes, while hard negative mining balances positive and negative anchors in each mini-batch, maintaining a 1/3 positive and 2/3 negative anchor ratio. Non-Max Suppression (NMS) filters overlapping predictions for the same object, prioritizing predictions at the center of a patch. Due to GPU memory limitations, cropped patches of CT scans are used as input for each training iteration. During inference, an overlapping sliding window is applied across the whole volume, with NMS handling overlapping predictions. Weighted Box Clustering is employed to consolidate predictions from multiple models or test-time augmentations. Training hyperparameters are detailed in Table~\ref{tab_sup_model_hyperparam}. To enhance the robustness of the model and improve generalization across input variability, the training applies data augmentation techniques, such as elastic deformation, scaling, rotation, mirroring along axes, and adjustments in brightness.

\subsubsection{Patch extraction and filter in lung module (CADe)} 
The full volume was resampled to a predefined spacing of $0.625\times0.625\times0.625$. Detected candidates were recomputed to match the new spacing, and candidates without any IoU overlap with the lung mask were filtered out. The conservative threshold of $0$ for IoU was chosen to minimize the risk of missing nodules located at the lung boundaries (e.g., pleura). A patch of size $64\times64\times64$ centered on each candidate was then extracted for segmentation and characterization.

\subsubsection{Nodule segmentation module (CADe)} 
The model is trained on CT scan crops of size $64\times64\times64$ with a fixed spacing of $0.625\times0.625\times0.625$ around the centers of the bounding boxes of the nodule detection. Since lesion segmentation is done on a 3D patch, some lesions may extend beyond the patch size. Only true positive detected boxes are chosen to train the model, which means each training sample contain a corresponding mask. Based on this training set, we trained ten separate models, deploying a 10-fold cross-validation strategy to improve the generalizability of the model. The same method is applied for the test set, which means only true detected boxes are used for evaluation. This network was based on a 3D-U-Net\cite{ronneberger_u-net_2015} architecture, which is able to process the entire 3D patch as input, and output a 3D mask of matching dimensions. We normalized HU intensity scale to the $[0,1]$ range, then the patch is rotated, and flipped. The addition of several noises (e.g., Gaussian and Gibbs noise) promotes robustness. Dice loss was used to optimize the model’s objective. All training hyperparameters are provided in Table~\ref{tab_sup_model_hyperparam}. The resulting masks are subsequently used for radiomic and 3D-morphomic computation, and for volume or diameter measurements.

\subsubsection{Diameter measurement module (CADe)} 
This module computes the diameter for each segmented mask obtained from the detection stage. The major axis is defined as the longest arc on the convex hull of the 2D mask taken on the slice exhibiting the largest area.
The minor axis is defined as the longest segment included in the convex hull of the mask orthogonal to the long axis on the same slice. In cases where the 3D segmentation mask comprises multiple connected components, only the largest component by volume is considered. The resulting diameter is used as an input filter for all subsequent (characterization) processes to exclude all detections that fall outside the $[4, 40]$mm range. This filtering step not only considerably reduces computation time and charge of the product, but also decreases FPs, thereby improving both characterization performance and training time.

\subsubsection{Nodule characterization module (CADx)} 
CT scans are first resampled to a uniform voxel dimension 0.625×0.625×0.625 (mm) using bicubic spline for the CT images and Nearest Neighbors for masks. A basic HU normalization is applied, consisting of a percentile crop (below 1\% and above 99\%) followed by a normalization to ensure that numerical values in the input data fall within the $[0,1]$ interval. Each of the 15 models is then trained on 15 different training subsets of Train3, each comprising all malignant findings and one of the 15 non-overlapping, randomly split and equally sized partitions of the benign findings. This procedure increased classifier diversity and considerably reduced training times, while improving the overall accuracy, by counterbalancing the very large imbalance of the training set. Each of the training sets was then further split into training (80\%) and tuning (20\%) sets.\\
\paragraph{3D Densenet module:} The 3D model takes as input a patch centred on a nodule and outputs a malignancy score for that nodule during both training and inference. For each detection, a 3D patch centered on the corresponding bounding box of dimensions $64\times64\times64$ is extracted. The model architecture is a shallow 3D DenseNet39 (with 39 layers) inspired by the work of Yuan \textit{et al.}\cite{yuan_effective_2019}. It consists of three Dense-S Blocks, each containing four Dense Conv Blocks, with a growth rate of 32 and dropout rate (p=1/2). The main difference between a 2D DenseNet\cite{huang_densely_2017-1} and its 3D counterpart lies in the replacement of 2D convolution and 2D pooling layers with their 3D equivalents. Basic data augmentations is applied at random during training, including flips and ’light’ affine transforms (with probability = 0.5, translate range = $(5,5,5)$ voxels, rotate range = $(2\pi,2\pi,2\pi)$, using reflection padding and excluding scaling transformations, as malignancy is not considered scale-invariant). As with 2D models, 3D models are trained using a focal loss function. Training hyperparameters are provided in Table~\ref{tab_sup_model_hyperparam}.\\
\paragraph{2D CNN module:} For each detected nodule, $64\times64$ patches were extracted from all slices containing target nodules (in axial view). The model consists of three convolutional layers, one max pooling layer and one mean pooling layer, producing a learned feature vector as output. A learned bilinear transform is applied between the feature vector of a given patch and the average feature vector of all patches in the lesion, in order to produce a contextualized feature vector. A final dense layer outputs a two-dimensional vector, where the second coordinate represents the malignancy likelihood of the input nodule. To break linearity, the network also includes ReLU layers, and to enforce and enhance generalizability, it uses dropout layers and batch normalization applied prior to each pooling layer. This model is trained using a focal loss function to address the high imbalance in the dataset. Nodule-wise inference is made by computing a malignancy score for each patch within the nodule and then taking the maximum score among them as the final output, yielding a single malignancy score for each nodule of a given patient. Training hyperparameters are given in TableS1.\\
\paragraph{XGBoost classifiers module:} Following~\cite{van_griethuysen_computational_2017,aerts_decoding_2014}, 111 radiomic features are extracted (see \href{https://pyradiomics.readthedocs.io/en/latest/features.html}{pyradiomics} for detailed description), including:
\begin{itemize}
    \item 17 Morphological features (margins, spiculation, sphericity): 3D shape-based (one additional redundant feature of "volume" computed by counting the voxels volumes), including multiple different diameters calculus, and nodule surface area, surface to volume ratio, sphericity among other  interesting features.
    \item 94 texture/luminance features (for attenuation, solid, part solid, ground glass, calcification): 19 first order Statistics of HU, 24 Gray Level Cooccurrence Matrix (GLCM), 16 Gray Level Run Length Matrix (GLRLM), 16 Gray Level Size Zone Matrix (GLSZM), 5 Neighboring Gray Tone Difference Matrix (NGTDM), 14 Gray Level Dependence Matrix (GLDM).
\end{itemize}
The Pyradiomics documentation details each radiomic feature, all computed in or from the mask of segmentation in the ($64\times64\times64$ voxels) 3D CT patches of the lesions (within the masks for the textural features). Additionally, our model extracts 11 3D-morphomics as described in~\cite{munoz_3d-morphomics_2022} that proved to be very efficient for malignancy classification. The resulting 122 3D-morphomics and radiomics are then used to train and test a gradient boosted decision tree classifier (XGBoost\cite{chen_xgboost_2016}, \href{https://github.com/dmlc/xgboost/tree/master}{DMLC XGBoost}).
Training: XGBoost hyper-parameters are tuned on the NLST train-tuning set with a 80-20\% split using Bayesian descent on a binary logistic loss (hyperopt library). The explored space and corresponding range is:'eta': loguniform $[1E-5,0.1]$; 'max depth': $[1,2,3,4,5]$; 'gamma': loguniform $[1E-2,7]$; 'alpha': loguniform $[1E-8,1E-2]$; 'lambda': loguniform $[1,4]$; 'colsample bytree': uniform $[0.5,0.7]$; 'colsample bylevel': uniform $[0.5,1]$; 'min child weight': qloguniform $[1,30,1]$; 'subsample': qloguniform $[0.1,0.7,0.05]$; 'scale pos weight': uniform $[1,\text{imbalance}]$; 'num boost round': quniform $[100,1000,30]$. The number of estimators is set to 180 with initialization at default XGboost hyperparameters. The final model ultimately used for inference is then re-trained on the full train set with optimal parameters with a learning rate of 0.01 and 1,000 estimators.
 
\subsubsection{Nodule predictions ensembling and calibration module (CADx)}  
As an output of the malignancy classifiers, we have for each lesion, 15 3D model predictions, 15 2D model predictions and 15 XGBoost predictions. Average prediction per model: The first step of the ensembling computes the mean of each class of models to obtain three mean predictions. Optimal convex domain ensembling: The second step consist in a stacking ensemble of the three predictions\cite{mohammed_comprehensive_2023} (convex domain affine optimization) using \href{https://hyperopt.github.io/hyperopt/}{Hyperopt} library trained on the training set, that derive the three optimal coefficients for those models reflecting each model’s predictive power: 
\begin{itemize}
  \item mean 3D-CNN=92.38\%; 
  \item mean 2D-CNN=1.84\%; 
  \item mean XGBoost=5.78\%. 
\end{itemize}  
  
The lesion level intermediate prediction of malignancy is then computed as the weighted sum of those three mean predictions.\\
\textbf{Calibration:} The final step involved calibrating the lesion level prediction using Netcal logistic calibration optimization model
by Küppers \textit{et al.}\cite{kuppers_multivariate_2020} (\href{https://github.com/EFS-OpenSource/calibration-framework}{GitHub repository}). Logistic calibration temperature scaling provided superior calibration performances. The calibrated ensembled prediction demonstrated significantly higher AUC than any of the three mean models alone.

\subsubsection{Full-CT scan characterization module (CADx)} 
This model is designed to capture contextual features at both patient level and at lesion level, e.g., emphysema, pleural wall thickness, lesion multiplicity, inflammations, mediastinal and hilar lesions (although the latter are beyond the scope of the present study). It aims at improving diagnostics accuracy by reducing patient and lesion level errors. Data preprocessing: The model operates on 3D chest CT scans. Each CT scan is first cropped around the center of the bounding box determined by its lung segmentation. The scans are then resampled to a uniform volume size of $256\times256\times200$ voxels by nearest neighbor down-sampling, preserving the original spacing along the z-axis by allowing for cropping or padding. The HU intensity values are cropped below -1020 and above 300, consistent with a standard lung windowing, then rescaled to the $[0,1]$ interval, excluding the lowest and highest 1\% of intensity values.
Model : The model we use in this version of the product is a variant of Sybil, developed by Mikhael \textit{et al.}\cite{mikhael_sybil_2023}. Initially designed to predict cumulative cancer risk over one to six years, we adapted the output of the model to fit a binary criterion (patient-level
“cancer” vs. “non-cancer”) at $T_{-1}$. The model relies solely on 3D cropped CT scans as input for inference and whereas during training it additionally uses 3D masks of detected nodules. This flexibility allows for the model to enforce attention to lesion localizations during training, thereby enhancing performance\cite{mikhael_sybil_2023}, while limiting inference costs.
Optimization: Training relies on a dual-term loss, combining a Focal loss term which controls the proximity of the predicted score to the binary GT, and an attention loss term ensuring localization of detected nodules. Unlike the original implementation, we remove the loss term predicting the side of the lung where the potential cancer has or will grow. The network is initialized with a Resnet-18 encoder pretrained on the Kinetics-400 dataset. Due to the model’s complexity relative to the number of data instances, affine augmentations are performed during training. Training hyperparameters are provided in Table~\ref{tab_sup_model_hyperparam}.

\subsubsection{Nodule prediction update and calibration module (CADx)}
The ensembling of scan-level prediction combines the output from five different elementary scan predictions : four full volume models prediction and one lesion based prediction (the maximum of the predictions over all detection per scan). Rescaling: As a preliminary preprocessing step, all the five predictions are linearly rescaled to the $[0,1]$ interval such that the minimum prediction becomes 0 and the maximum 1. The parameters governing these linear transformations are obtained on the training set, and stored in our model as learned parameters. Optimal convex domain ensembling: mirroring the nodule level, the second step employs a stacking ensemble using the same method, to derive the five optimal coefficients for those models representing their respective predictive power:
\begin{itemize}
    \item Nodule (max) model = 24.89\%
    \item Full-CT scan model 3 = 23.10\%
    \item Full-CT scan model 1 = 21.41\%
    \item Full-CT scan model 2 = 16.36\%
    \item Full-CT scan model 4 = 14.24\%
\end{itemize}
The intermediate scan level prediction is calculated as the weighted sum of those five predictions. Calibration: As for nodule level, the final step consisted in calibrating the ensembled patient prediction using exactly the same methodology. This ensembled and calibrated scan prediction provided significantly higher AUC than any of the five elementary model alone.
Nodule prediction update: This last step refines nodule-level malignancy prediction by adjusting them based on the final patient-level prediction. Such correction accounts for broader contextual cancer cues (emphysema...) that are beyond the scope of lesion level prediction which infer malignancy from only a very small and localized region of the image. The correction is computed by the algorithm as the delta between between the final patient prediction and the maximum lesion-level prediction for a patient, expressed as a percentage of the maximum lesion: \[\Delta \text{ correction} = \frac{\text{Final scan prediction} - \text{Max nodule prediction}}{\text{Max nodule prediction}}\]  
The final updated lesion prediction was then obtained as:  
\[\text{Final nodule prediction} = \text{Intermediate nodule prediction} + \Delta \text{ correction} \times \text{Intermediate nodule prediction}.\]   This update-correction slightly improves lesion-level performance compared to nodule-level prediction alone.

\subsubsection{Multiple detections reduction module and diameter measurement (CADx)}
Owing to the detection model output, identical finding may occasionally be detected multiple times across multiple patches. The deduplication eliminates such redundant detections through a two steps process:
\begin{itemize}
    \item \textit{Clean finding segmentations:} (1) remove connected components with diameter $<4$ mm; (2) keep only the connected component closest to the patch center, removing others. Patches were centered on the detected bounding box center. Ensuring connected components were centered in the patch guaranteed recovery of all detected findings.  
    \item \textit{Merge cleaned findings:} (1) group findings whose masks intersected; (2) merge finding groups into their union; and (3) assign the highest malignancy score among merged findings to the resulting finding.
\end{itemize}
This procedure guarantees that segmentations of reported findings do not overlap with one another, and that the merged finding receives the maximal malignancy prediction. For pairing purposes, the detection bounding box of the merged finding is set to the bounding box of the merged
segmentation. The segmentation bounding box is preferable for pairing since the segmentation may contain multiple findings
detected independently with smaller detection boxes. The updated diameter extraction method operates on deduplicated findings: the diameter extraction method used following the nodule segmentation module remains unchanged. Then the SAX is extracted directly from the segmentation, guaranteeing
it lies on the segmentation border. Whether the segmentation itself or its convex hull is used does not affect the extraction of the long axis (LAX), which remains unchanged. This new constraint—that the extremities of the SAX must lie on the segmentation contour—, produces the following edge case: the SAX may be too short to intersect with the LAX. In such instances, the SAX segment is extended until it intersects with the LAX. Additionally, differing from the previous method for diameter extraction and for aesthetic considerations: LAX and SAX are extracted from the largest 2D component on the slice with the greatest area.

\section*{Supplementary References}

\end{document}